\documentclass[10pt,twocolumn,letterpaper]{article}

\usepackage{iccv}
\usepackage{times}
\usepackage{epsfig}
\usepackage{graphicx}
\usepackage{amsmath}
\usepackage{amssymb}
\usepackage{subfigure}
\usepackage{xcolor}

\newtheorem{definition}{Definition}[section]
\newtheorem{observation}{Observation}[section]
\usepackage{booktabs}

\usepackage[accsupp]{axessibility}

\usepackage[breaklinks=true,bookmarks=false,colorlinks]{hyperref}
\iccvfinalcopy

\def\iccvPaperID{9471} 

\ificcvfinal\fi
\usepackage{multirow}
\begin{document}

\title{Feature Proliferation --- the ``Cancer'' in StyleGAN and its Treatments}

\author{Shuang Song \quad Yuanbang Liang \quad Jing Wu \quad Yu-Kun Lai \quad Yipeng Qin\thanks{Corresponding author: Yipeng Qin}\\[0.1cm]
Cardiff University\\
{\tt\small \{songs7,liangy32,wuj11,laiy4,qiny16\}@cardiff.ac.uk}
}

\maketitle
\ificcvfinal\thispagestyle{empty}\fi
\begin{abstract}
Despite the success of StyleGAN in image synthesis, the images it synthesizes are not always perfect and the well-known truncation trick has become a standard post-processing technique for StyleGAN to synthesize high quality images.
Although effective, it has long been noted that the truncation trick tends to reduce the diversity of synthesized images and unnecessarily sacrifices many distinct image features. 
To address this issue, in this paper, we first delve into the StyleGAN image synthesis mechanism and discover an important phenomenon, namely \textit{Feature Proliferation}, which demonstrates how specific features reproduce with forward propagation.
Then, we show how the occurrence of Feature Proliferation results in StyleGAN image artifacts.
As an analogy, we refer to it as the ``cancer'' in StyleGAN from its proliferating and malignant nature.
Finally, we propose a novel feature rescaling method that identifies and modulates risky features to mitigate feature proliferation.
Thanks to our discovery of Feature Proliferation, the proposed feature rescaling method is less destructive and retains more useful image features than the truncation trick, as it is more fine-grained and works in a lower-level feature space rather than a high-level latent space.
Experimental results justify the validity of our claims and the effectiveness of the proposed feature rescaling method. 
Our code is available at \url{https://github.com/songc42/Feature-proliferation}.
\end{abstract}

\section{Introduction}
Deep learning has entered the era of large models.
For instance, the GPT-3~\cite{brown2020language} developed by OpenAI has 175 billion parameters and can easily cost around 15 million dollars to train for a single run, let alone the GPT-3.5 underpinning the recently hyped ChatGPT\footnote{\url{https://openai.com/blog/chatgpt}}; Stability.ai trained their Stable-diffusion-v1-4\footnote{\url{https://huggingface.co/CompVis/stable-diffusion-v1-4}} using 256 Nvidia A100 GPUs for 150,000 GPU hours; Nvidia spent 92 GPU years and 225 MWh of electricity with an in-house cluster of NVIDIA V100 GPUs to develop StyleGAN3~\cite{karras2021alias} and 4 weeks on 64 NVIDIA
A100s for a ``constrained'' training of their recent StyleGAN-T~\cite{sauer2023stylegan}.
As a result, although the performance of large models is impressive,
their high costs have become a critical concern, {\it e.g.}, OpenAI admitted that there was a bug in their GPT-3 model but cannot afford to retrain it due to the high cost~\cite{brown2020language}.
This motivates us to do our best to avoid retraining large neural network models.
In this work, we focus on the StyleGAN family~\cite{karras2019style,karras2020analyzing,karras2021alias,sauer2023stylegan} as they are inherently more efficient (\ie, generating images with a single pass), allow for excellent semantic interpolation in their latent spaces, and are comparable to the quality of diffusion models~\cite{sauer2023stylegan}.
For the StyleGAN series, although there were no obvious bugs in the training, the synthesized images are not always perfect.
While instead of attempting to solve this problem by improving the model design which requires multiple retraining, StyleGAN follows the ``no retraining'' philosophy and resorts to a post-processing technique called truncation trick~\cite{marchesi2017megapixel,brock2018large}.
In short, the truncation trick improves the quality of StyleGAN synthesized images by normalizing their corresponding latent codes towards their mean in the {\it latent} space.
However, despite its popularity, it has long been noted that the truncation trick tends to reduce
the diversity of synthesized images and unnecessarily sacrifices many distinct image features~\cite{karras2019style}.

\begin{figure*}
  \centering
  \includegraphics [width=0.98\linewidth]{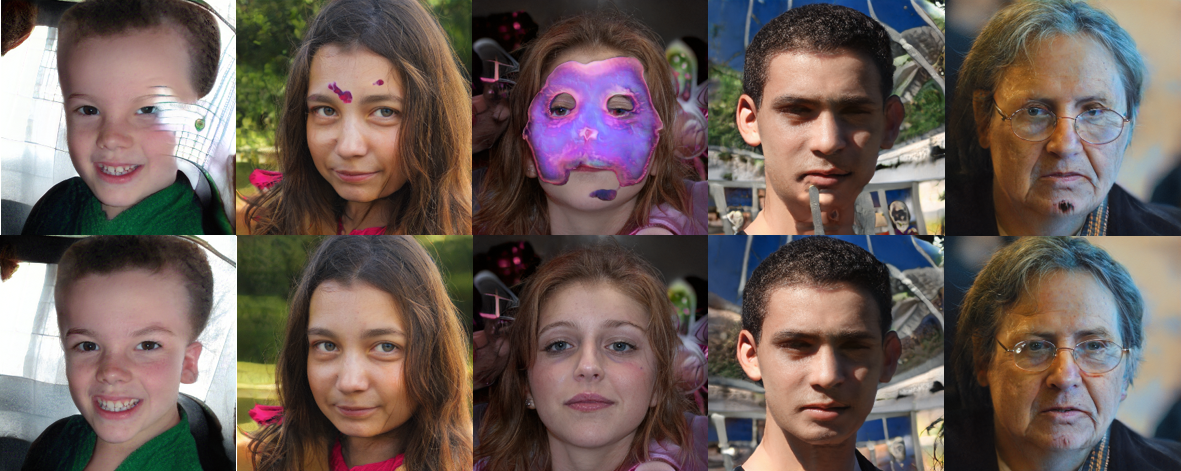}
  \vspace{1mm}
  \caption{Top: StyleGAN synthesized images with artifacts. Bottom: images ``cured'' by our method.}
  \vspace{-2mm}
  \label{Fig_1}
\end{figure*}

In this paper, we address the low-quality images synthesized by StyleGAN from a different perspective, {\it i.e.}, its mechanism for image synthesis in the {\it feature} space.
Specifically, we delved into the architectural details of the StyleGAN generator and discovered an important phenomenon, namely {\it Feature Proliferation}, which demonstrates how specific features reproduce with forward propagation.
In short, Feature Proliferation denotes the phenomenon where the ratio of the occurrence of a certain type of feature (those with abnormally large values) in a layer increases with forward propagation.
Our analysis points out that such a phenomenon is a by-product of the weight modulation and demodulation techniques used in StyleGAN2/3~\cite{karras2020analyzing,karras2021alias} and the latest StyleGAN-T~\cite{sauer2023stylegan}.
Interestingly, we observed that Feature Proliferation usually leads to artifacts in StyleGAN synthesized images and these artifacts can be easily removed by a simple feature rescaling method.
To minimize the unnecessary interference with useful image features, we propose a novel method to identify the risky features prone to proliferation in the earliest layers, thereby removing the least amount of them.
As a result, our method significantly outperforms the truncation trick in terms of retaining useful image features while improving the quality of StyleGAN synthesized images.
Experimental results justify the validity of our claims and the effectiveness of the proposed feature rescaling method.
Our contributions include:
\begin{itemize}
    \item We discover an important phenomenon, namely {\it Feature Proliferation}, which shows how specific features reproduce with forward propagation. We also show that it is a by-product of the weight modulation and demodulation techniques used in modern StyleGANs.
    \item We discover a strong causal relationship between the occurrence of Feature Proliferation and StyleGAN image artifacts.
    \item We propose a novel feature rescaling method that identifies and modulates risky features to mitigate feature proliferation, thereby achieving a better trade-off between quality and diversity of StyleGAN synthesized images than the popular truncation trick. 
\end{itemize}

\section{Related work}
\label{gen_inst}
Since the seminal work of Goodfellow et al.~\cite{goodfellow2014generative}, Generative Adversarial Networks (GANs) have become one of the most promising deep generative models that have numerous applications in computer vision and graphics.
Nevertheless, GANs are known to be notoriously difficult to train and a variety of techniques have been proposed to stabilize their training from different perspectives, including architectures~\cite{radford2015unsupervised,he2016deep}, loss functions~\cite{arjovsky2017wasserstein,mao2017least}, regularization~\cite{mescheder2018training,miyato2018spectral}, and the interactions between them~\cite{qin2020does}.
Thanks to the massive efforts from the community, the StyleGAN series~\cite{karras2019style,karras2020analyzing,karras2021alias,sauer2023stylegan} have become one of the most influential models in the GAN family, as it can generate high-resolution and realistic images that are almost indistinguishable from real photos by human inspectors.
Examples of its applications include GAN inversion and editing~\cite{abdal2019image2stylegan,abdal2020image2stylegan++,wang2021high,alaluf2022hyperstyle}, image-to-image translation~\cite{richardson2021encoding}, super-resolution~\cite{menon2020pulse}, 3D shape generation~\cite{zheng2022sdf}, etc.
We refer interested audiences to~\cite{bermano2022state} for a survey on StyleGAN and its applications.
Nevertheless, despite its success, the quality of images synthesized by StyleGAN varies, and a post-processing method known as the truncation trick~\cite{marchesi2017megapixel,brock2018large} has widely been adopted to obtain higher quality images.
However, the truncation trick has long been noted as ``destructive'' as it unnecessarily sacrifices useful image features and tends to reduce the diversity of synthesized images.
In this paper, we address this issue by investigating the StyleGAN image synthesis mechanism and propose a feature rescaling method that can precisely identify the risky features causing image artifacts, thereby improving image quality by modulating these features at the minimal cost.

\section{Feature Proliferation in StyleGAN}
\label{SEC:Feature Proliferation}
In this section, we first introduce two new phenomena that we discovered in the forward propagation of deep neural networks, {\it i.e.} feature domination and proliferation. We then show that they are by-products of the weight modulation and demodulation techniques used in modern StyleGANs. Finally, we show how they affect StyleGAN and lead to artifacts in synthesized images.

\begin{figure}[t]
  \centering
  \includegraphics [width=0.95\linewidth]{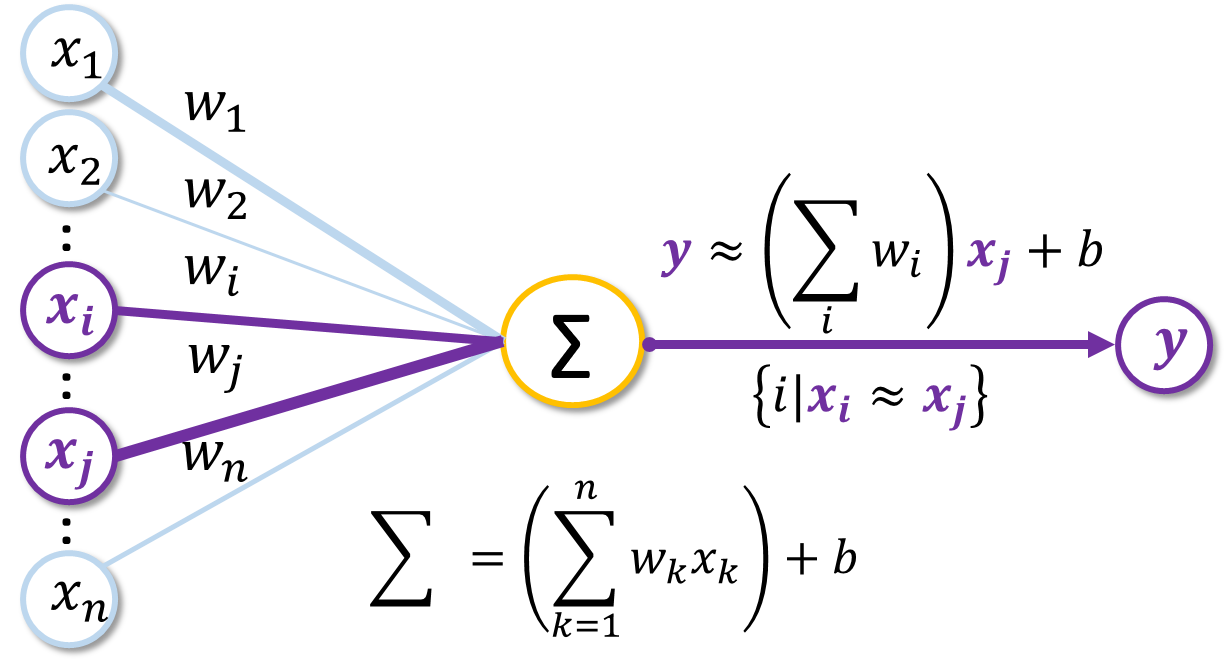}
  \caption{Illustration of feature domination. $y$ is dominated by $x_j$ and its perturbations $x_i \approx x_j$.}
  \label{fig:feature_domination}
\end{figure}

\subsection{Feature Domination and Proliferation}
{\raggedright Throughout the history of machine learning, the  weighted sum method has been built to last from the classical linear regression, to the theoretically elegant support vector machines and the deep neural networks that are now taking both academia and industry by storm. 
For deep neural networks, we usually refer to the output of neurons as {\it features} that are the elements to be weighted and summed.
However, due to the differences among weights and input features, the output of a neuron can occasionally be dominated by one or a small amount of input features, which leads to a phenomenon we call {\it Feature Domination}.}

\begin{definition}[Feature Domination]
Let $y = \mathbf{w^{T}}\mathbf{x} + b$ be the output of a neuron, $\mathbf{x}$ and $\mathbf{w}$ are the input and weight vectors respectively, $b$ denotes the bias, we define {\it Feature Domination} as the case where $y \approx (\sum_{i}w_i)x_j + b$ is dominated by feature $x_j$ and its perturbations $x_i \approx x_j$, where $i$ indicates the $i$-th element in $\mathbf{w}$ and $\mathbf{x}$.
\label{def:feature_domination}
\end{definition}

In a nutshell, the proposed {\it Feature Domination} describes the phenomenon that one or a small number of similar input features may dominate the weighted sum if their products with associated weights significantly outweigh others (Fig.~\ref{fig:feature_domination}), making $y$ similar to $x_i$/$x_j$.
In practice, we observed that the dominant features are highly likely to be those that deviate significantly from the mean of their distributions:

\begin{observation}[Feature Domination in StyleGAN]
 Let $t$ be an empirically obtained threshold, and  $x_{l,j}$ be the feature map of channel $j$ of layer $l$ of the generator. $x_{l,j}$ is highly likely to be dominant if $\frac{|\mathrm{mean}(x_{l,j}) - \mu_{x_{l,j}}|}{\sigma_{x_{l,j}}} > t$, where $\mathrm{mean}(x_{l,j})$ is the mean value of all elements in $x_{l,j}$, $\mu_{x_{l,j}}$ and $\sigma_{x_{l,j}}$ are the mean and standard deviation of the distribution of $x_{l,j}$ over the training dataset.
\label{ob:feature_domination}
\end{observation}

In practice, we use Observation~\ref{ob:feature_domination} to identify dominant features in StyleGANs.
Furthermore, {\it Feature Proliferation} happens when the same type of Feature Domination proliferates during the forward propagation.

\begin{definition}[Feature Proliferation]
Following Definition \ref{def:feature_domination}, let $\mathbf{Y}_d(x_j)$ be the set consisting of all outputs $y$ dominated by feature $x_j$ in a single neural network layer, we define {\it Feature Proliferation} as the increase of $\eta(x_j)=|\mathbf{Y}_d(x_j)|/|\mathbf{Y}|$ during forward propagation, where $\mathbf{Y}$ is the set of all outputs in a single neural network layer and $|\cdot|$ denotes the cardinality of a set. 
\label{def:feature_proliferation}
\end{definition}

When {\it Feature Proliferation} happens, one or a small number of features will dominate the output of the entire network, leading to artifacts in StyleGAN synthesized images.
As an analogy, we refer to it as the ``{\bf cancer}'' in StyleGAN from its proliferating and malignant nature. 

\vspace{2mm}
\noindent \textbf{Remark.}
For simplicity, Definitions~\ref{def:feature_domination} and~\ref{def:feature_proliferation} are introduced with fully-connected layers. However, both definitions can be easily extended to other types of neural networks, {\it e.g.}, convolutional neural networks (CNNs)~\cite{wu2017introduction}, as the weighted sum operation is widely used in almost all deep neural networks due to its benefits in parallelization.

\begin{figure}[t]
\centering
\subfigure[Mean of inputs]{
\begin{minipage}[t]{0.5\linewidth}
\centering
\includegraphics[width=1\linewidth]{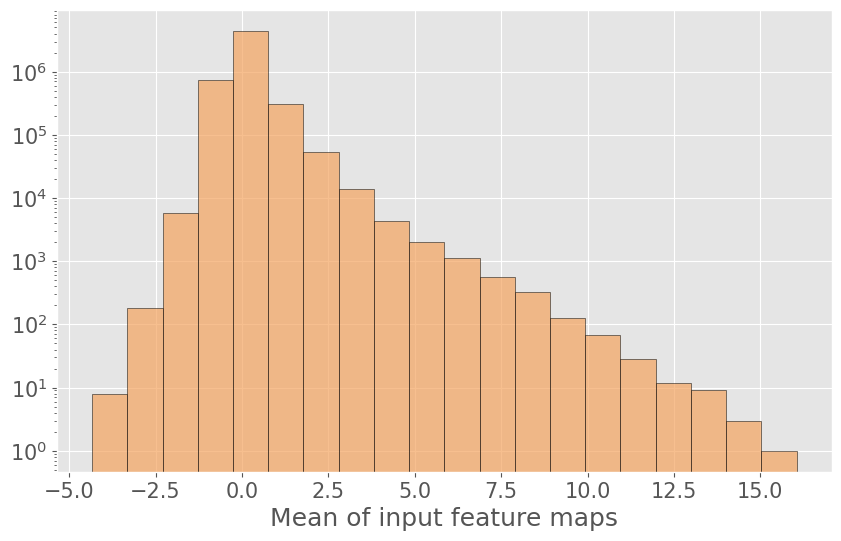}
\end{minipage}
}%
\subfigure[Standard deviation of inputs]{
\begin{minipage}[t]{0.5\linewidth}
\centering
\includegraphics[width=1\linewidth]{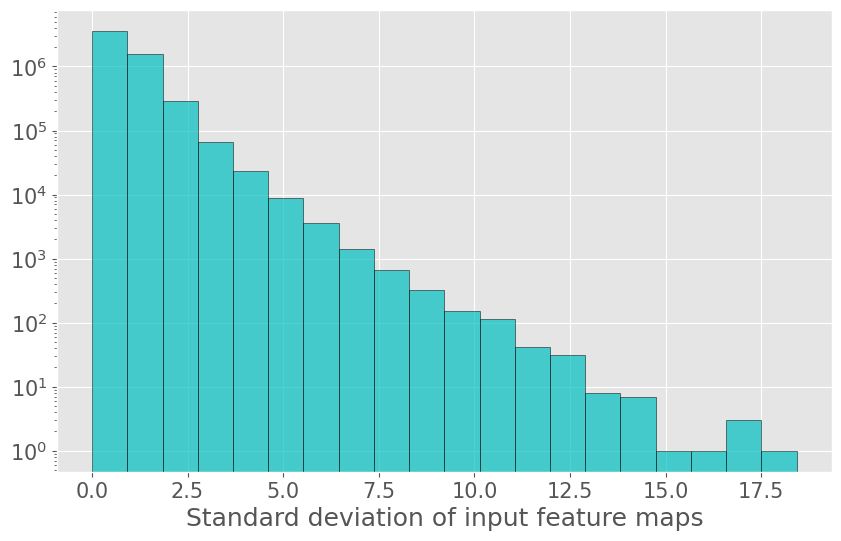}
\end{minipage}
}%

\caption{Histograms of mean and standard deviation of the input feature maps from 10,000 StyleGAN2 synthesized images.}
\vspace{-4mm}
\label{fig:N_0_I}
\end{figure}

\begin{figure*}
  \centering
  \includegraphics[width=0.98\textwidth]{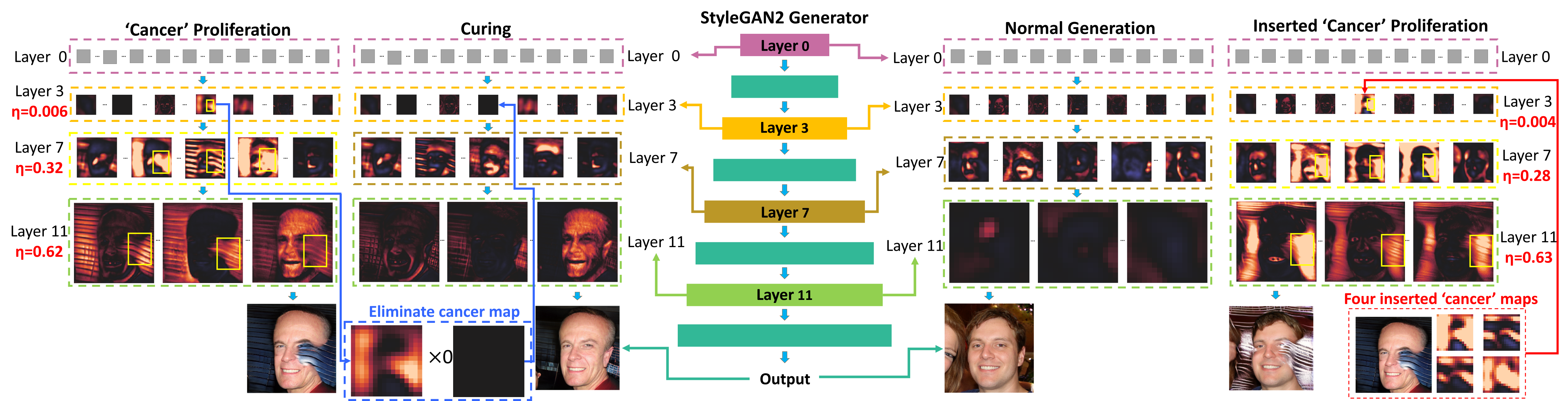}\label{Section3_v6}
  \vspace{4mm}
    \caption{Left: ('Cancer' Proliferation) the Feature Proliferation phenomenon leads to image artifacts; (Curing) setting the visually identified proliferating feature map in Layer 3 to zero (\textcolor{blue}{blue} arrow) mitigates feature proliferation. Right: (Normal Generation) a proper StyleGAN image synthesis process; (Inserted 'Cancer' Proliferation) inserting the ``cancer'' features identified in the left part results in feature proliferation and similar image artifacts (\textcolor{red}{red} arrow). The \textcolor{yellow}{yellow} boxes in feature maps highlight the feature proliferation process. $\eta$ in \textcolor{red}{red}: the ratio of dominated features in a layer.}
  \label{Section3}
\end{figure*}

\subsection{Root Causes of Feature Proliferation}
\label{sec:root_cause}

We ascribe feature proliferation to the strong statistical assumption used in the weight modulation and demodulation of StyleGAN2/3/T \cite{karras2020analyzing,karras2021alias,sauer2023stylegan},
\ie, ``the input activations are {\it i.i.d.} random variables with unit standard deviation'', rather than the statistics (\ie, mean and standard deviation) of the actual feature maps used in the AdaIN of StyleGAN1~\cite{karras2019style}.

Specifically, weight demodulation assumes that all input activations (modulated feature maps) are already normalized to be of unit standard deviation, thereby eliminating the step of feature map normalization $\frac{\mathbf{x}_{i}-\mu\left(\mathbf{x}_{i}\right)}{\sigma\left(\mathbf{x}_{i}\right)}$ used in AdaIN.
Since the standard deviations of output feature maps are determined by the products of those of their input feature maps (assumed to be unit ones as mentioned above) and convolutional weights, the abovementioned assumption ensures the unit standard deviations of output feature maps by only ``demodulating'' the convolutional weights to be of unit standard deviations~\cite{karras2020analyzing}:
\begin{equation}
w_{i j k}^{\prime \prime}=w_{i j k}^{\prime} / \sqrt{\sum_{i} w_{i j k}^{\prime 2}+\epsilon}, w_{i j k}^{\prime}=s_{i} \cdot w_{i j k}
\end{equation}
where $s_i$ is the scaling parameter corresponding to the $i$-th input feature map, $j$ and $k$ enumerate the output feature maps and spatial footprint of the convolution, respectively.
While in practice, the assumption of the unit standard deviation of input activations does not hold (Fig.~\ref{fig:N_0_I}). As a result, weight demodulation cannot ensure the unit standard deviations of output feature maps.
Instead, the standard deviations of output feature maps are determined by those of input ones, causing feature domination and proliferation.

\vspace{2mm}
\noindent \textbf{Remark.}
AdaIN explicitly normalizes each feature map according to its mean and standard deviation so the mean and standard deviation of its output are solely determined by the style parameters $\mathbf{y}=\left(\mathbf{y}_{s}, \mathbf{y}_{b}\right)$ but not those of its input feature map $\mathbf{x}_{i}$~\cite{karras2019style}:
\begin{equation}
\operatorname{AdaIN}\left(\mathbf{x}_{i}, \mathbf{y}\right)=\mathbf{y}_{s, i} \frac{\mathbf{x}_{i}-\mu\left(\mathbf{x}_{i}\right)}{\sigma\left(\mathbf{x}_{i}\right)}+\mathbf{y}_{b, i}
\end{equation}
This breaks the chain of feature proliferation in forward propagation and we did not observe significant proliferation in our experiments. However, as shown in~\cite{karras2020analyzing}, AdaIN has its own problems (\ie, the ``characteristic artifacts'', which is why weight demodulation is used in StyleGAN2/3) and may not be a good solution to feature proliferation.

\subsection{Impact of Feature Proliferation on StyleGAN}
\label{sec:impact}

As Fig.~\ref{Section3} shows, the synthesized images suffer from obvious artifacts when Feature Proliferation happens and we observed a strong spatial correlation between the proliferated feature maps and the artifacts in synthesized images. 
To justify the causality between them, we show that i) the artifacts can be ``cured'' by removing corresponding features before their proliferation in Fig.~\ref{Section3} (left) and ii) the artifacts can be added to high-quality images by inducing feature proliferation in Fig.~\ref{Section3} (right).

\begin{figure*}
  \centering
  \subfigure[StyleGAN2 pretrained with FFHQ dataset]{
  \includegraphics[width=0.48\textwidth]{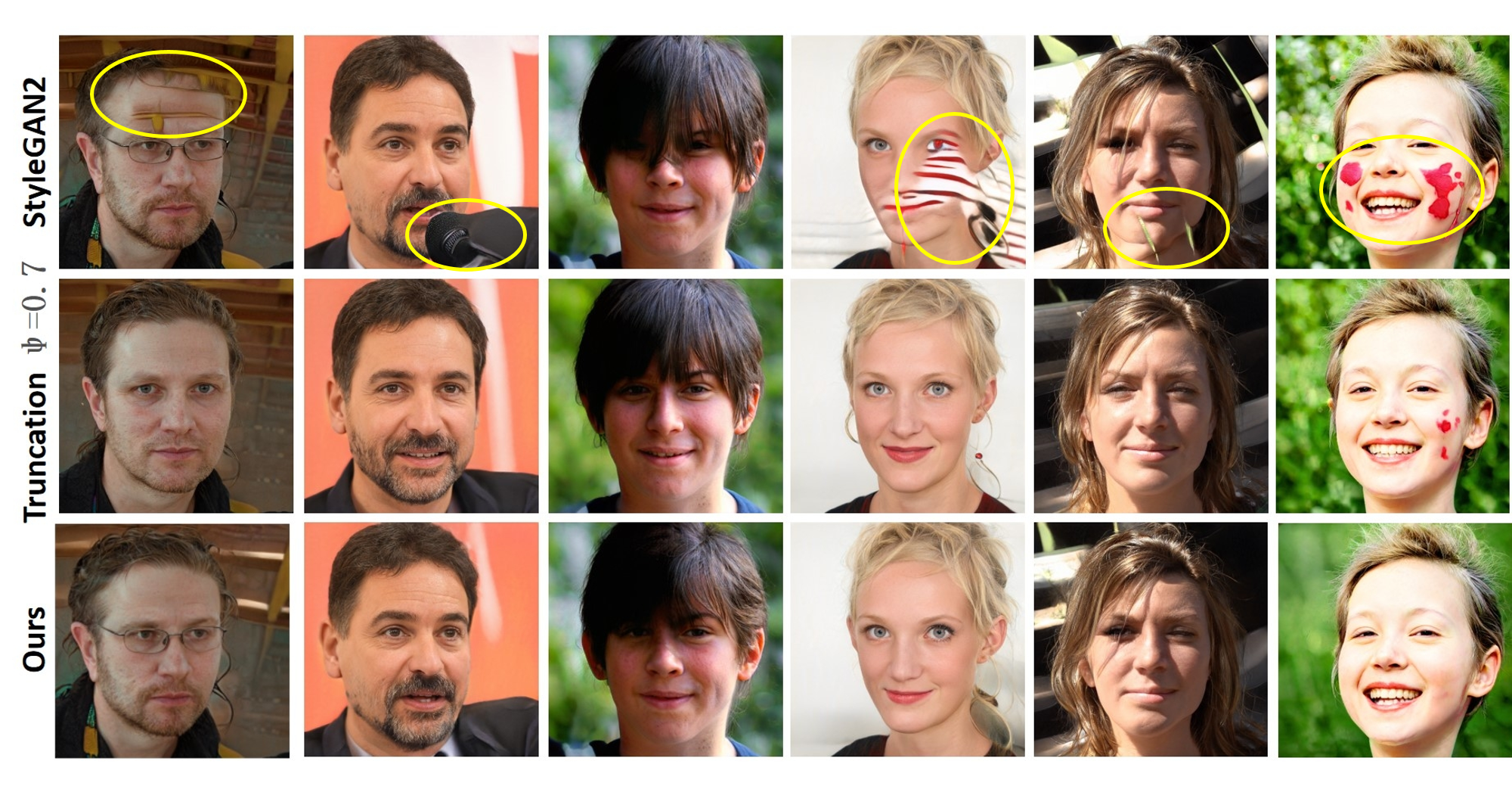}\label{FFHQ}
  }
  \subfigure[StyleGAN2 pretrained with AFHQ-Cat dataset]{
  \includegraphics[width=0.48\textwidth]{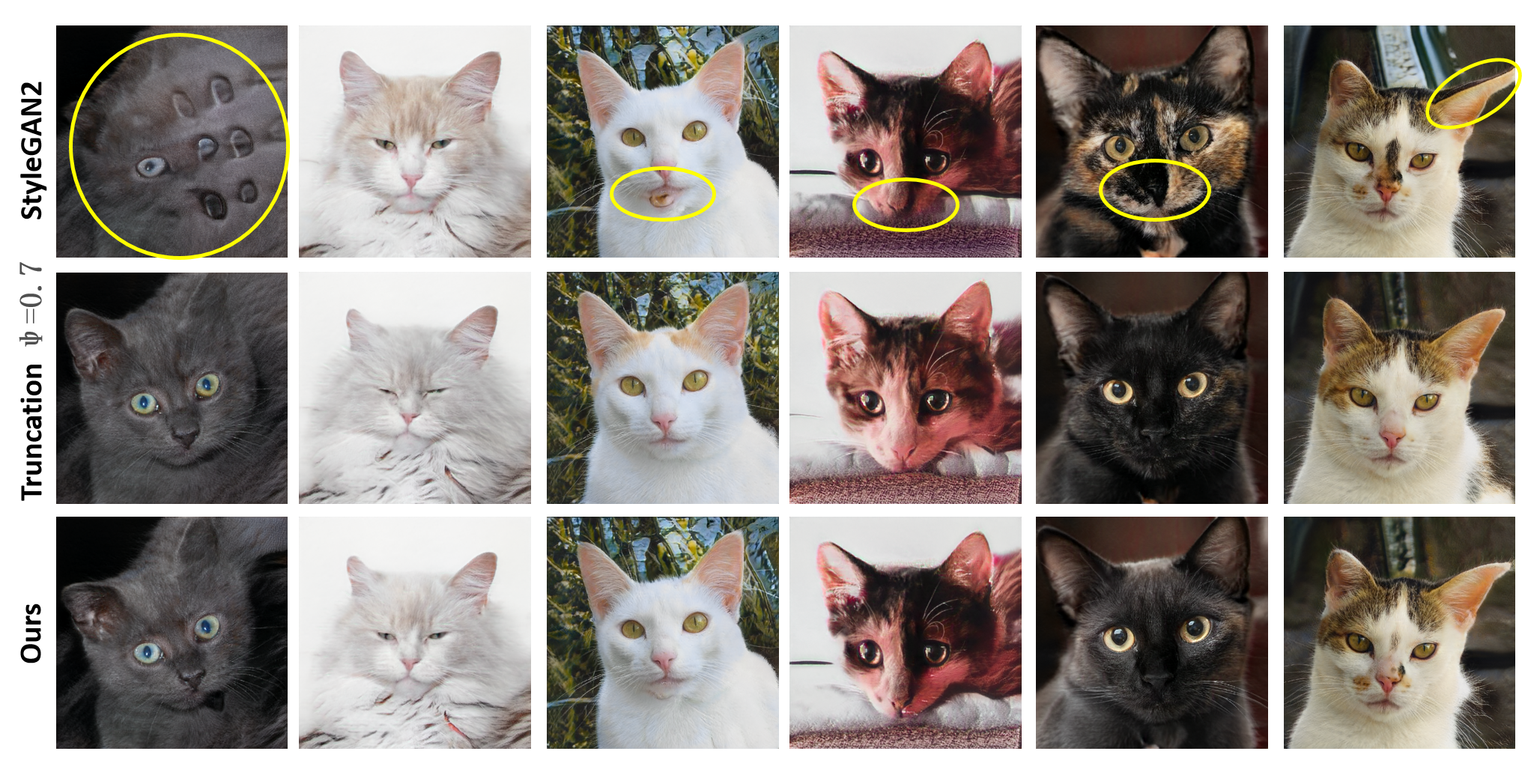}\label{FHAQ_cat}
  }
  \subfigure[StyleGAN3 pretrained on FFHQ dataset]{
  \centering
  \includegraphics[width=0.48\textwidth]{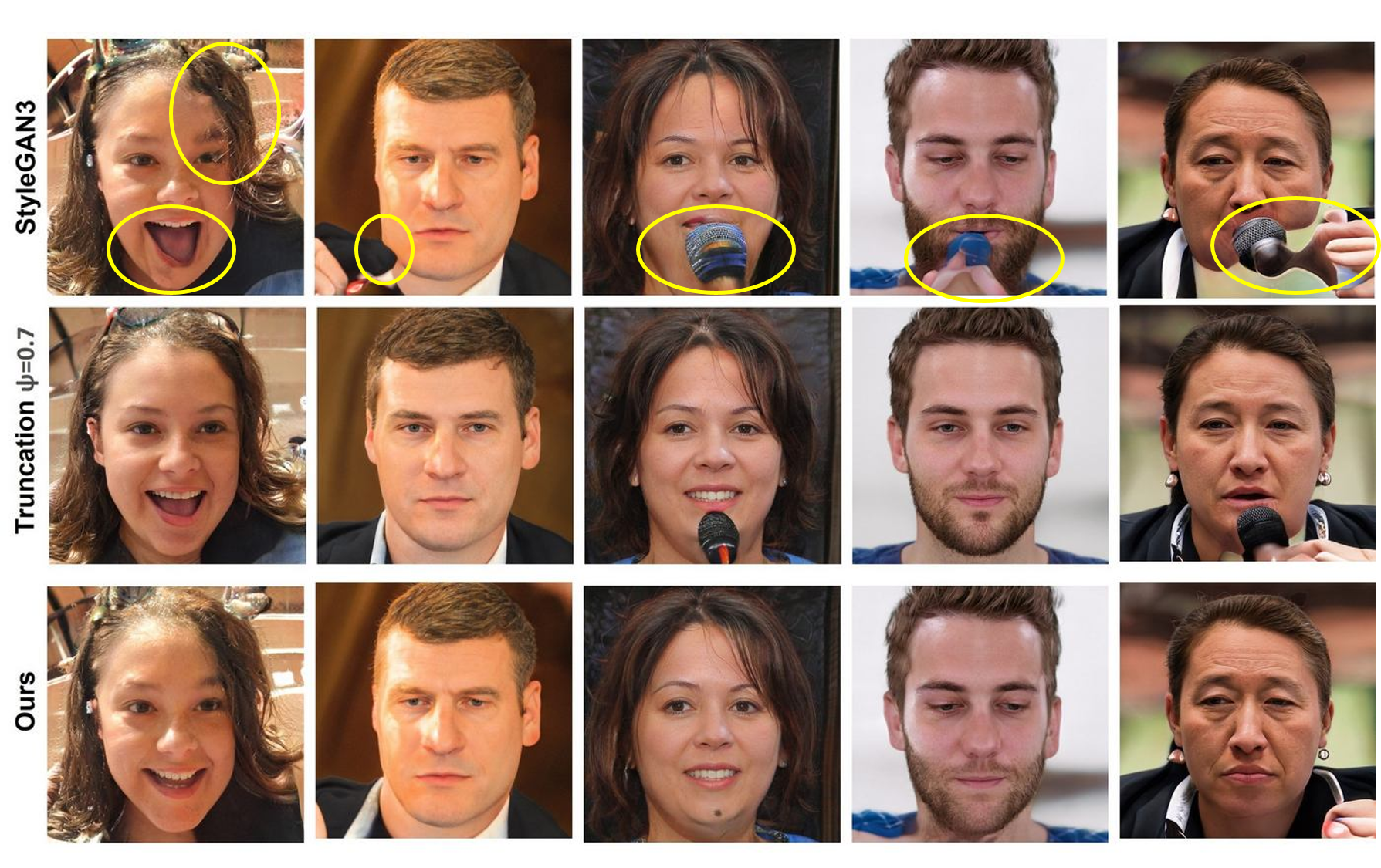}\label{SG3_FFHQ}
  }
  \subfigure[StyleGAN3 pretrained on AFHQ dataset]{

  \centering
  \includegraphics[width=0.48\textwidth]{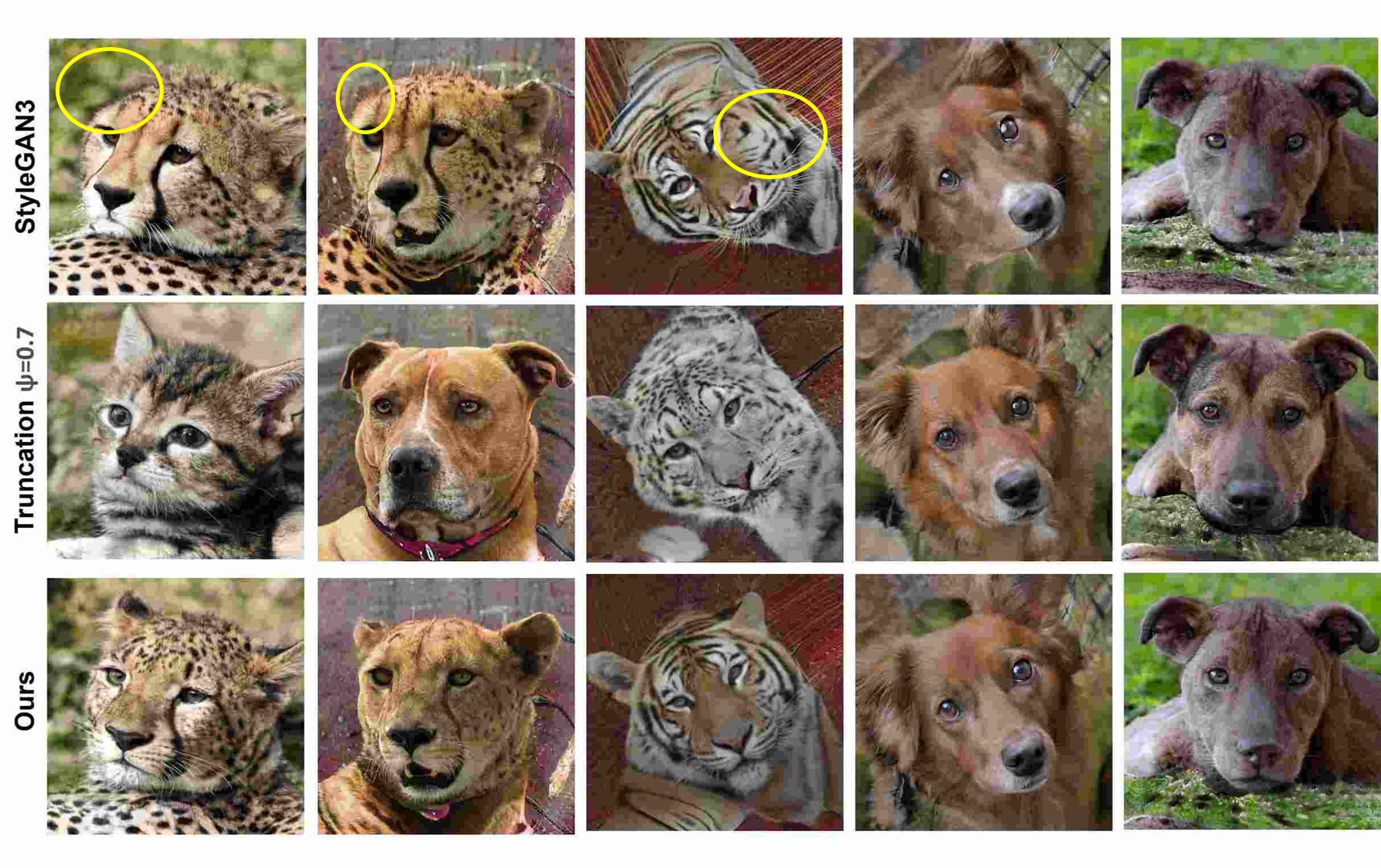}\label{SG3_AFHQ}
  }
  \vspace{4mm}
    \caption{Curated StyleGAN2/3~\cite{karras2020analyzing,karras2021alias} synthesized images, those processed by the truncation trick ($\psi$=0.7), and our method. }
  \label{Fig5_1}
\end{figure*}

\section{Curing the ``Cancer'' in StyleGAN}\label{sec4}

In this section, we first introduce how to identify the risky features for feature proliferation and then introduce a simple but effective feature rescaling method to adjust the identified features and mitigate the feature proliferation.

\subsection{Feature Proliferation (``Cancer'') Identification}
\label{SUBSEC:Feature Proliferation (``Cancer'') Identification}
Naively, one of the most straightforward methods for feature proliferation identification is to extract all similar pairs of feature maps in the successive layers of a neural network. 
However, let $k$ be the number of successive pairs of layers and $n$ be the number of neurons per layer, the time complexity of this strategy is $O(kn^2)$ per image, which is inefficient and makes it infeasible for frequent use.

Addressing this issue, we employ a heuristic stemming from Observation~\ref{ob:feature_domination} that the dominant features are likely to be those that deviate significantly from the mean of their distributions. 
Specifically, we estimate the mean $\mu_{l,j}$ and standard deviation $\sigma_{l,j}$ with a large number ({\it e.g.}, 3,000) of randomly sampled images and identify $x_{l,j}$ as a high risk feature for proliferation if: 
\begin{equation}
r_{l,j}=\frac{\left|\mathrm{mean} (x_{l,j})-\mu_{l,j}\right|}{\max({\sigma_{l,j},c})}> t \label{eq:identification}
\end{equation}
where $\mathrm{mean} (x_{l,j})$ is the mean value of all elements in $x_{l,j}$, $r_{l,j}$ denotes the amount of risk of feature proliferation, $t$ is a user-defined threshold obtained empirically, $c=0.1$ is a small constant introduced to avoid the mis-identification when $\sigma_{l,j}$ is small.

\subsection{Curing the ``Cancer'' by Feature Rescaling}

Since the risky features identified by Eq.~\ref{eq:identification} proliferate with forward propagation, we propose a simple yet effective feature scaling method to address the proliferation in the earliest layers, thereby removing the least amount of them and minimizing the unnecessary interference with useful image features. Let $x_{c}$ be a ``cancer'' feature map with risk $r$ (Eq.~\ref{eq:identification}), we have
\begin{equation}
x_{c}^{m}=\frac{x_{c}}{p\cdot r}\label{fea_mod}
\end{equation}
where $x_{c}^{m}$ denotes the modified feature map, $p$ is the scaling hyper-parameter. 
By rescaling $x_{c}$, we reduce the values of its elements and thus prevent it from Feature Domination (Definition~\ref{def:feature_domination}) and Proliferation (Definition~\ref{def:feature_proliferation}), thereby improving the quality of StyleGAN synthesized images.

\section{Experiments}

Please see the {\bf supplement} for more experiments.

\subsection{Experimental Setup}

The proposed method does not require network training.
We run the proposed method with pretrained deep generative models on a workstation with an Intel(R) Core(TM) i7-10875H
CPU and a GeForce RTX 3080 GPU.
The pretrained StyleGAN2~\cite{karras2020analyzing} models (using FFHQ~\cite{karras2019style}, MetFace~\cite{karras2020training} and AFHQ-Cat~\cite{choi2020stargan} datasets) used in all our experiments are publicly-released on Github\footnote{
StyleGAN2: \url{https://github.com/NVlabs/stylegan2}}.
Unless specified, we use hyperparameters $t=2$ and $p=2$ for our method, and $\psi=0.7$ for the truncation trick that has been suggested for the best trade-off between quality and diversity of synthesized images~\cite{karras2019style,brock2018large}.

\subsection{Qualitative Evaluation}
As Fig.~\ref{Fig5_1} shows, we compare the performance of our method against the truncation trick~\cite{marchesi2017megapixel,brock2018large} on StyleGAN2 and StyleGAN3 models pretrained on FFHQ, MetFace and AFHQ-Cat datasets.
In Fig.~\ref{FFHQ}, it can be observed that: i) all the StyleGAN2 synthesized images in the top row, except the third one, contain obvious artifacts; ii) for the third column (high-quality image), our method retains useful image features better than the truncation trick; iii) for the other columns, our method not only removes the artifacts but also retains useful image features in a better way, {\it e.g.}, the truncation trick eliminates the eyeglasses in column 1 while our method retains them and other fine facial details like the beard.
Similar results can be observed in Fig.~\ref{FHAQ_cat}, which demonstrates the versatility of our method across datasets.

Figs.~\ref{SG3_FFHQ} and \ref{SG3_AFHQ} show synthesized results of the truncation trick and our method on StyleGAN3~\cite{karras2021alias} models pretrained on FFHQ and AFHQ datasets\footnote{\url{https://github.com/NVlabs/stylegan3}}. It can be observed that our claims still hold on StyleGAN3, which further justifies the superiority of our method.

\begin{figure}[t]
  \centering
  \includegraphics [width=0.47\textwidth]{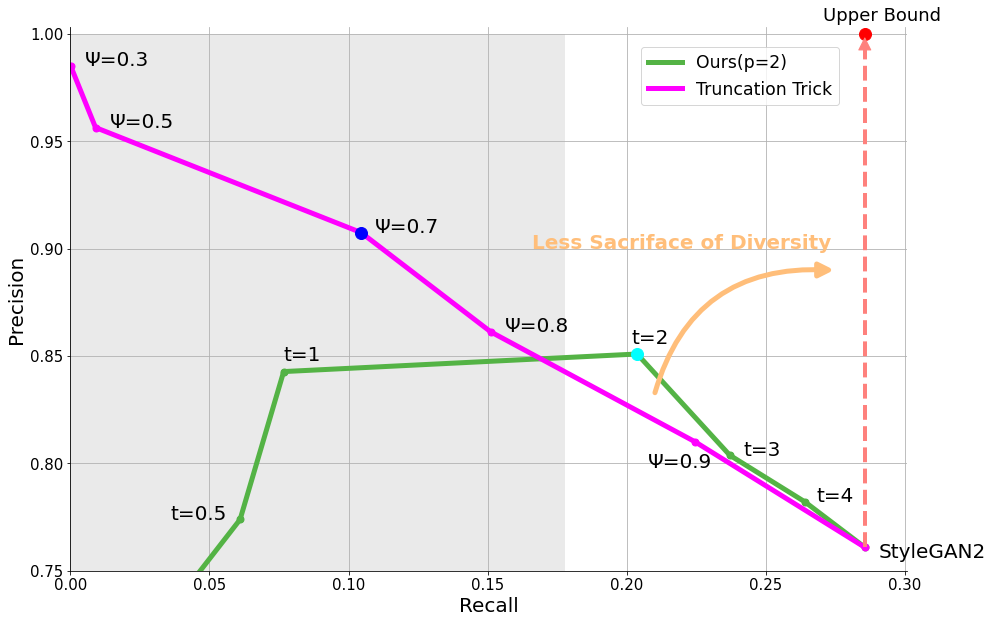}
  \caption{Comparison of precision and recall~\cite{kynkaanniemi2019improved} between the truncation trick~\cite{marchesi2017megapixel,brock2018large} and our method on StyleGAN2. $p$ and $t$ are hyperparameters of our method and $\psi$ is the hyperparameter for the truncation trick.
  ``StyleGAN2'' (bottom-right corner): precision and recall of the original StyleGAN2;
  \textcolor{blue}{Blue} ($\psi=0.7$): best $\psi$ recommended by StyleGAN2~\cite{karras2020analyzing}; \textcolor{cyan}{Cyan} ($p=2$, $t=2$): best $p$ and $t$ of our method;
  \textcolor{red}{Red}: the upper bound of StyleGAN ``correction'' methods where every error is corrected (precision=1.0) without sacrificing diversity (recall remains constant);
  \textcolor{orange}{Orange}: the closer to the dashed line, the less sacrifice of diversity;
  \textcolor{gray}{Gray}: region of {\it uninterest} where too much diversity has been sacrificed.
  Ours is closer to the dashed line in the region of interest, indicating that it strikes a better balance between precision and recall.
  }
  \label{fig:P_R_FFHQ}
\end{figure}

\begin{table}
  \caption{Comparison with Truncation Trick (TT)~\cite{brock2018large} using PSNR, SSIM~\cite{ssim}, LPIPS~\cite{lpips}, ID (identity preservation  using the ArcFace~\cite{deng2019arcface}) and FID~\cite{fid} computed with 10K images on StyleGAN2.}
  \label{tb:PSNR_SSIM_LPIPS}
  \centering
  \begin{tabular}{lrrrrrr}
    \toprule
    Dataset
    \centering
     &\multicolumn{2}{c}{AFHQ-Cat}  &\multicolumn{2}{c}{MetFace} &\multicolumn{2}{c}{FFHQ}                     \\
    Method & TT& Ours & TT& Ours& TT & Ours
     \\
    \midrule
    PSNR$\uparrow$&3.66 &\textbf{4.11} & 5.33& \textbf{10.45}& 3.80& \textbf{4.28}\\
    SSIM$\uparrow$&0.79  & \textbf{0.81}  & 0.81& \textbf{0.87}   & \textbf{0.74}  &0.68  \\
    LPIPS$\downarrow$& 0.15 & \textbf{0.14}  &0.22  & \textbf{0.14}  &0.31 & \textbf{0.24} \\
    ID$\uparrow$& 0.74 & \textbf{0.84}  &0.77  & \textbf{0.87}  &0.71 & \textbf{0.77} \\
    FID$\downarrow$ & 7.43 & \textbf{4.47} & 23.02 & \textbf{17.51} & 12.56 & \textbf{11.91} \\
    \bottomrule
  \end{tabular}
\end{table}

\begin{figure}[t]
  \centering
  \includegraphics [width=0.99\linewidth]{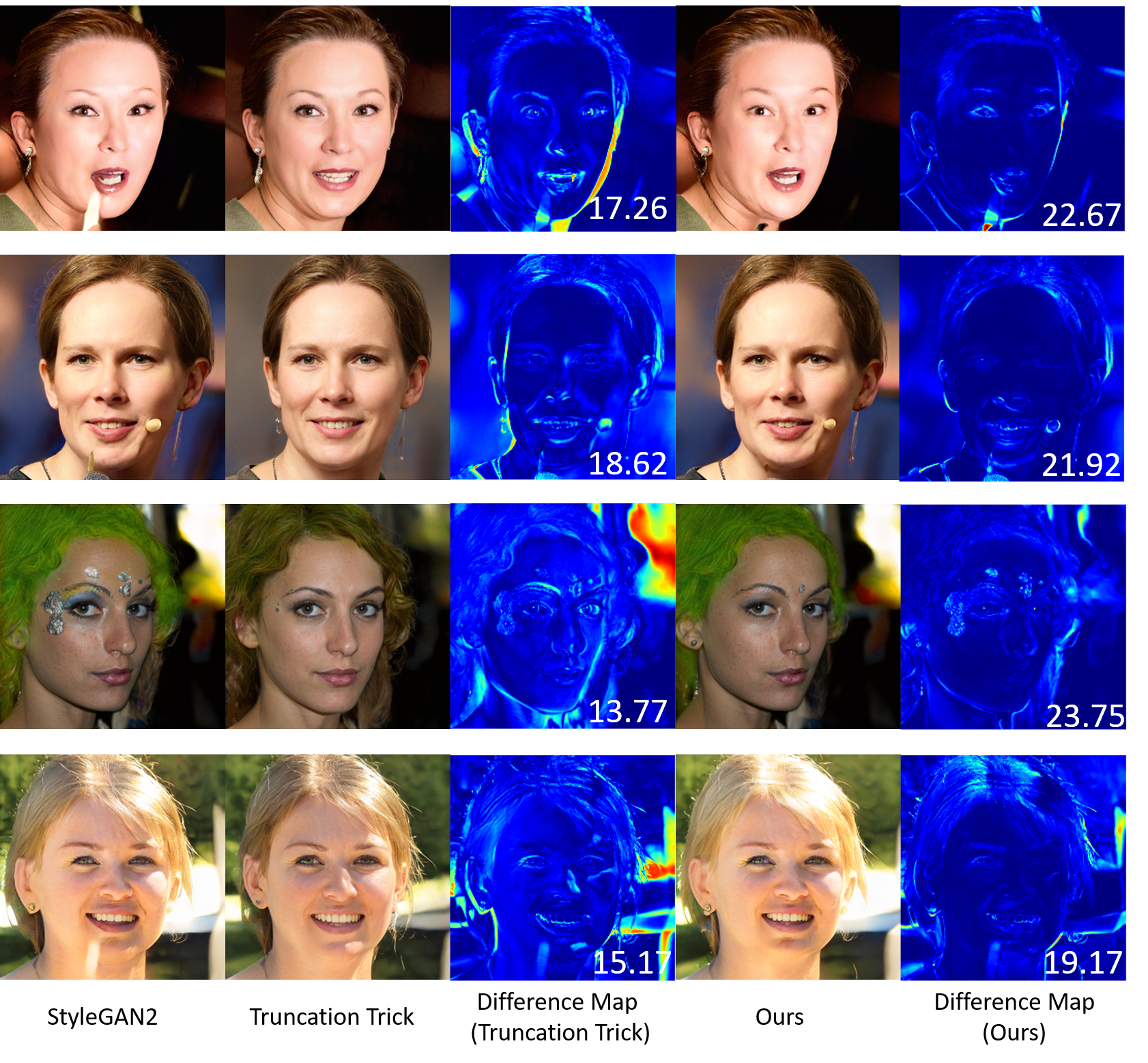}
    \caption{Comparison of difference maps of truncation trick~\cite{marchesi2017megapixel,brock2018large} and our method. The numbers at the bottom right corner of the difference images are PSNR scores.}
\vspace{-2mm}
  \label{fig:difference_map}
\end{figure}

\begin{figure*}
  \centering
  
  \begin{minipage}[t]{1\linewidth}
  \centering
  \includegraphics[width=0.99\linewidth]{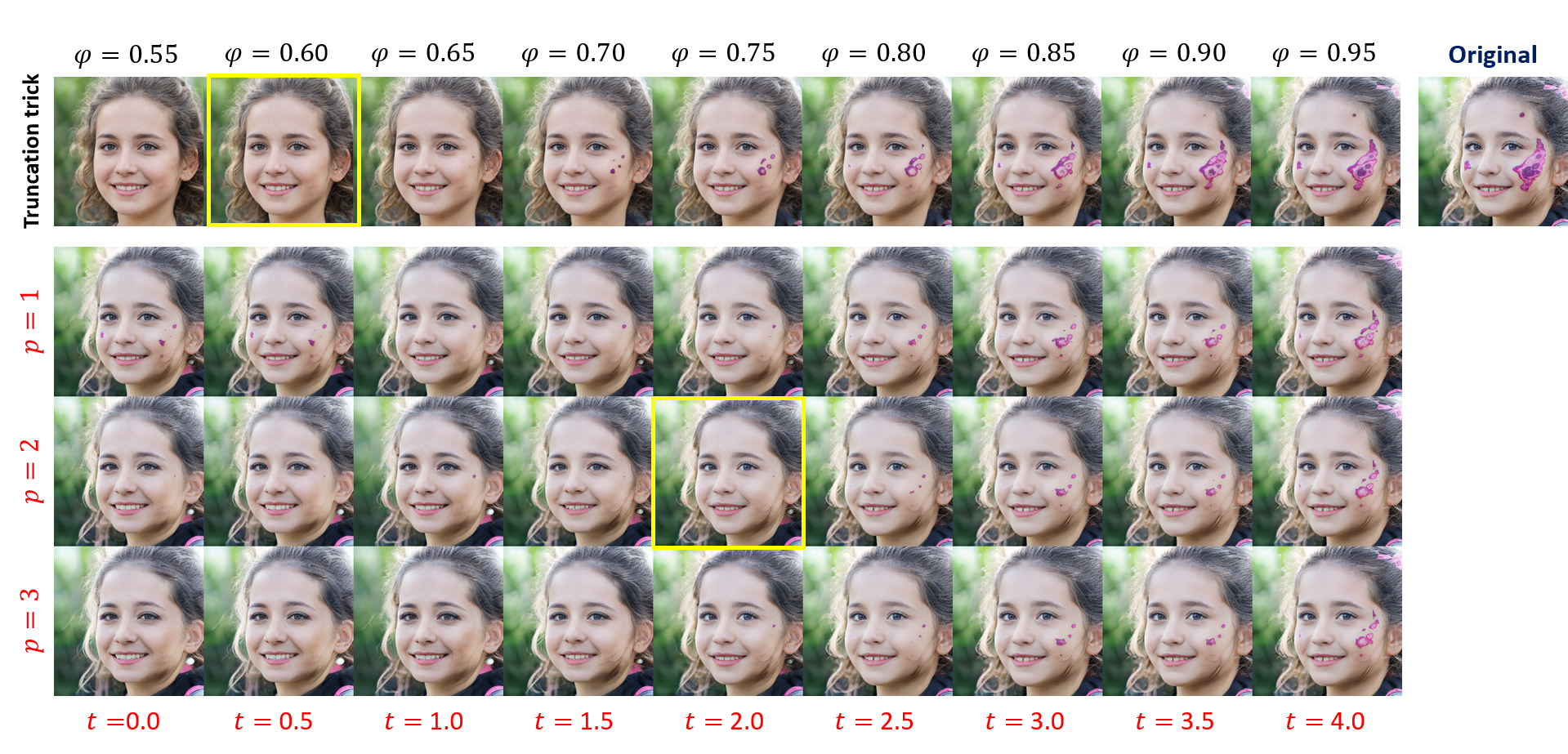}\label{hyper_t_p_1}
\end{minipage}%
\vspace{-2mm}  
\caption{Choice of hyper-parameters. Top row: images generated with truncation trick with $\psi=0.55$ to $0.95$. Rows 2-4: images generated with our method with $t=0.0$ to $4.0$, $p=1$ to $3$. Our method removes the artifacts while retaining almost all important features of the original image, indicating that it achieves a better trade-off between image quality and diversity than the truncation trick.}
  \label{hyper}
\vspace{-4mm}
\end{figure*}

\begin{table}
  \caption{Comparison with Truncation Trick (TT)~\cite{brock2018large} using PSNR, SSIM~\cite{ssim}, LPIPS~\cite{lpips} and FID~\cite{fid} computed with 10K images on StyleGAN3.}
  \label{SG3:PSNR_SSIM_LPIPS}
  \centering
  \begin{tabular}{lrrrr}
    \toprule
    Dataset
    \centering
     &\multicolumn{2}{c}{FFHQ}  &\multicolumn{2}{c}{AFHQ-Cat}                     \\
    Method & TT& Ours& TT & Ours
     \\
    \midrule
    PSNR$\uparrow$&16.23 &\textbf{18.07} & 16.03& \textbf{17.72}\\
    SSIM$\uparrow$&0.70  & \textbf{0.78}     &0.70 & \textbf{0.76} \\
    LPIPS$\downarrow$& 0.33 & \textbf{0.22}  &0.16 & \textbf{0.13} \\
    FID$\downarrow$ & 22.45 & \textbf{14.31} & 24.02 &\textbf{17.36} \\
    \bottomrule
  \end{tabular}
\end{table}

\subsection{Quantitative Evaluation}

\noindent \textbf{Precision and Recall.}
Fig.~\ref{fig:P_R_FFHQ} shows the comparison of precision and recall~\cite{kynkaanniemi2019improved} between the truncation trick~\cite{marchesi2017megapixel,brock2018large} and our method on StyleGAN2.
It can be observed that our method strikes a better balance between precision and recall ($p=2, t=2$) in the region of interest (right side, white region) while the truncation trick sacrifices too much diversity to achieve the same level of precision.
Recognizing that the aims of StyleGAN correction methods are to maximize image quality (\ie, precision) without sacrificing diversity (\ie, recall), our method has made a concrete step towards the ultimate solution (\textcolor{red}{red} point in Fig.~\ref{fig:P_R_FFHQ}).

\vspace{2mm}
\noindent \textbf{Image Similarity Metrics.}
Although less effective in assessing the trade-off between quality and diversity, for the sake of completeness, we also quantitatively compare the extent to which image features are retained by our method and the truncation trick~\cite{marchesi2017megapixel,brock2018large}.
As Table~\ref{tb:PSNR_SSIM_LPIPS} shows, for StyleGAN2, it can be observed that i) for PSNR, SSIM~\cite{ssim} and LPIPS~\cite{lpips}, our method outperforms the truncation trick in most cases, which demonstrates that our method better retains the useful features in the original image. However, the SSIM of our method is slightly worse than that of the truncation trick for the FFHQ dataset. We ascribe this to the superior power of our method in removing structural artifacts that are common in StyleGAN pretrained with the FFHQ dataset.
ii) For ID, we used the ArcFace~\cite{deng2019arcface} to compute the identity similarity scores between the original and the processed images.
Interestingly, we observe that although trained on human face datasets, ArcFace~\cite{deng2019arcface} generalizes to AFHQ-Cat~\cite{choi2020stargan} and MetFace~\cite{karras2020training} images and provides meaningful scores.
iii) For FID~\cite{fid}, our method outperforms the truncation trick in all three datasets, which again demonstrates that our method better retains the useful features in the original image.
We also show the difference maps in Fig.~\ref{fig:difference_map} to facilitate intuitive understanding.
The results on StyleGAN3~\cite{karras2021alias} are shown in Table~\ref{SG3:PSNR_SSIM_LPIPS}. It can be observed that our claims still hold on StyleGAN3, which further justifies the superiority of our method.

\subsection{Choice of Hyper-parameters}

Similar to $\psi$ in the truncation trick~\cite{marchesi2017megapixel,brock2018large}, we use two hyperparameters $t$ and $p$ for the trade-off between quality and diversity in our method.
Between them, $t$ determines how many feature maps are identified as risky ones for feature proliferation, and $p$ adjusts the extent to which we rescale such feature maps.
In general, the more risky feature maps identified, the stronger rescaling, and the higher quality but less diversity of synthesized images.

\vspace{1mm}
\noindent \textbf{Qualitative Justification.} As Fig.~\ref{hyper} shows, we visualize the results generated with the truncation trick with $\psi=0.6$ to $1.0$ and our method with $t=0.0$ to $4.0$ and $p=1.0$ to $3.0$, and identify the best choice ($p=2$, $t=2$) through visual inspection. 
In general, the smaller the $t$, the more features are identified as risky (Eq.~\ref{eq:identification}) for rescaling; the larger the $p$, the stronger the rescaling.
In this way, $t$ and $p$ work together to modify only the problematic features, thus achieving a better trade-off between quality and diversity than the truncation trick.

\begin{figure}[t]
  \centering
  \includegraphics [width=0.98\linewidth]{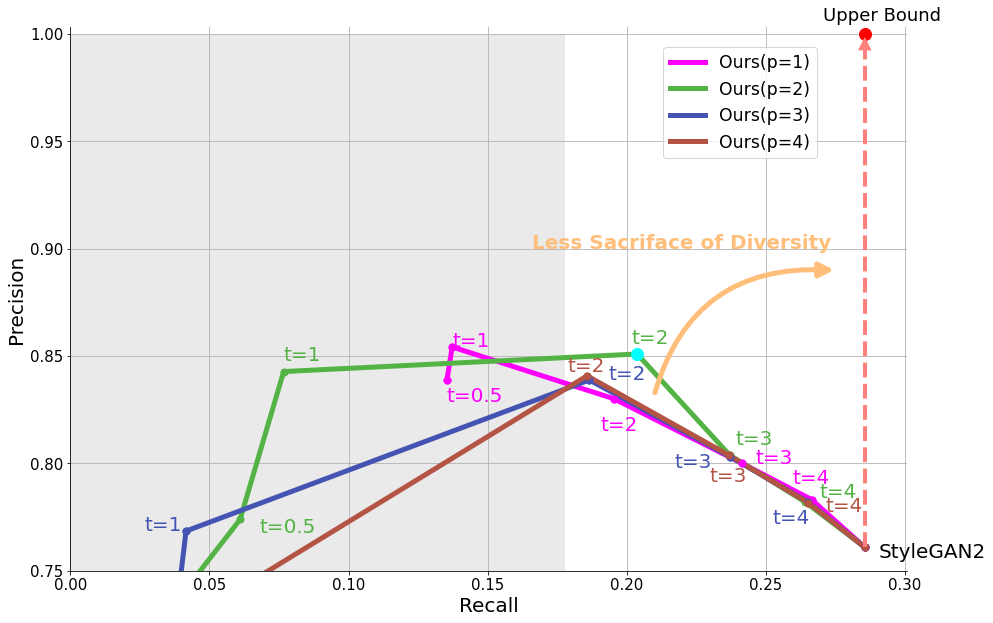}
  \vspace{1mm}
  \caption{Choice of Hyper-parameters (Quantitative). It can be observed that $p=2, t=2$ (\textcolor{cyan}{cyan}) achieves the best trade-off between precision and recall, which is consistent with the visual inspection results in Fig. \ref{hyper}.}
  \vspace{-4mm}
  \label{fig:addi_P_R}
\end{figure}

\vspace{1mm}
\noindent \textbf{Quantitative Justification.} We quantitatively justify the choice of hyper-parameters $p$ and $t$ by comparing their precision and recall.
As Fig.~\ref{fig:addi_P_R} shows, $p=2, t=2$ achieves the best trade-off between precision and recall, which is consistent with the qualitative results in Fig. \ref{hyper}.

\subsection{Ablation Study}

To justify the effectiveness of our channel-wise feature identification and rescaling strategy (Sec.~\ref{sec4}), we demonstrate its superiority over its layer-wise and pixel-wise variants as follows.

\begin{figure}[t]
  \centering
  \includegraphics[width=0.48\textwidth]{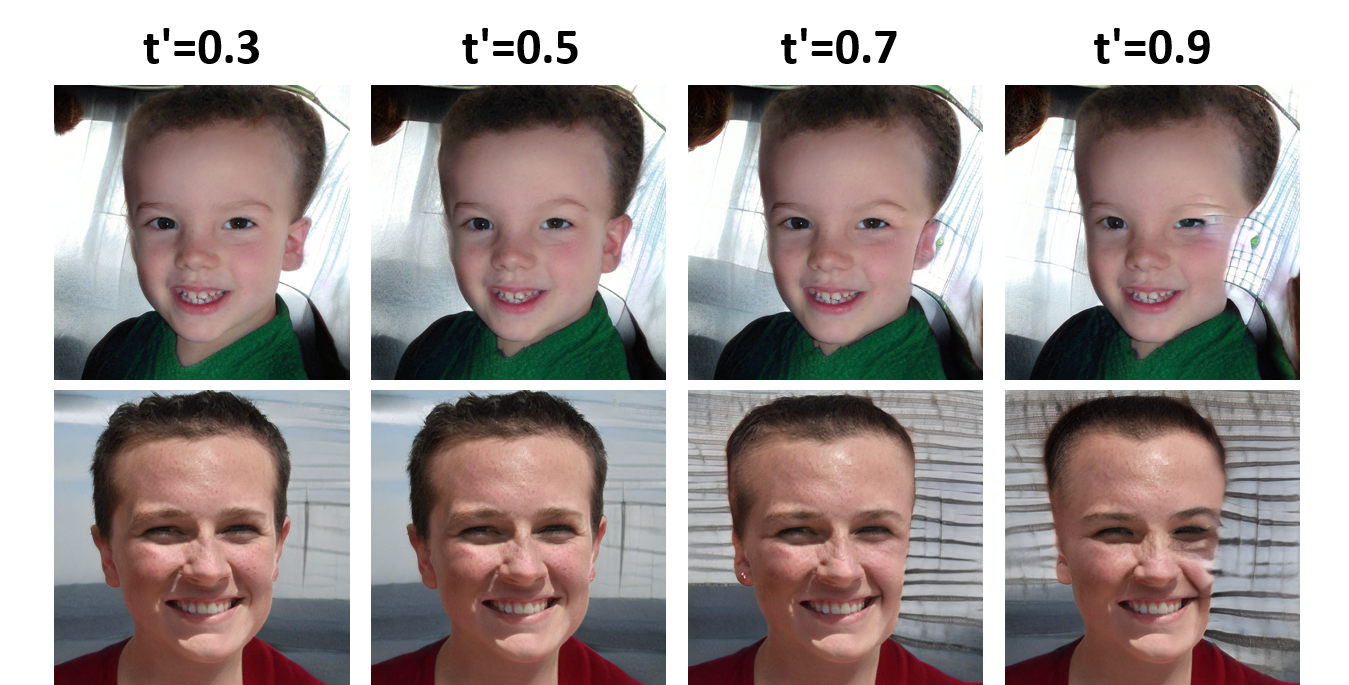}\label{ablation}
  \caption{Successful cases of the layer-wise variant of our method (threshold $t'$ ranges from 0.3 to 0.9). It can be observed that this variant is effective in removing artifacts of background intrusion.}
  \label{fig:ablation}
\end{figure}

\begin{figure}[h]
  \centering
  \includegraphics [width=0.49\textwidth]{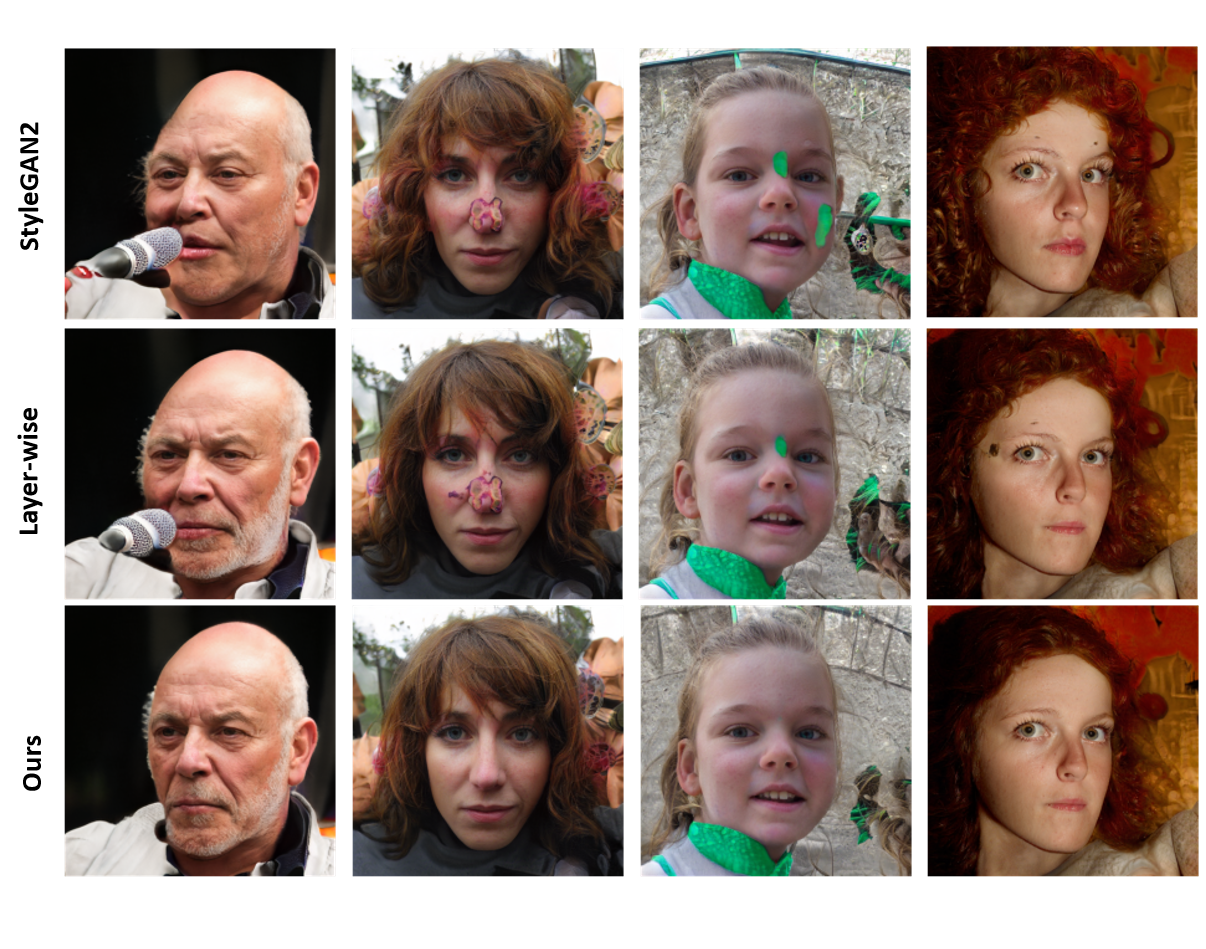}
    \caption{Failure cases of the layer-wise variant of our method (threshold $t'=0.5$). It can be observed that this variant is ineffective in removing artifacts of alien objects (\eg, color blocks on faces) while our method can remove them effectively. Top row: low-quality images synthesized by StyleGAN2; Middle row: results of the layer-wise variant of our method; Bottom row: results of our method.}

  \label{fig:Layer_wise}
\end{figure}

\vspace{1mm}
\noindent \textbf{Layer-wise ``Cancer'' Curing.} As in Sec.~\ref{sec4}, let $x_{l,j}$ be a ``cancer'' feature map identified using Eq.~\ref{eq:identification}, we compute the mean absolute ``cancer'' feature map for layer $l$ as:
\begin{equation}
\hat{x}_{l}=\mathrm{mean}_j(|x_{l,j}|)\label{eq:avera}\\
\end{equation}
where we take the absolute value $|x_{l,j}|$ to capture their magnitude and avoid them canceling each other out when added together.
In order to make the same choice of hyper-parameters applicable to all layers, we normalize $\hat{x}_{l}$ to $\bar{x}_{l}$ with mean $\mu=0$ and $\sigma=1$:
\begin{equation}
\bar{x}_{l}={\frac{\hat{x}_{l}-\mu{(\hat{x}_{l})}}{\sigma{(\hat{x}_{l})}}} \label{eq:average}
\end{equation}
Then we measure the ``correlation'' between each $|x_{l,j}|$ and its corresponding $\bar{x}_{l}$ as:
\begin{equation}
c_{l,j}=\frac{|x_{l,j}|*\bar{x}_{l}}{\sum_{\mathrm{element}}|x_{l,j}|} \label{eq:cof}
\end{equation}
where ${\sum_{\mathrm{element}}|x_{l,j}|}$ is the sum for all elements in $|x_{l,j}|$.
Let $t'$ be a threshold hyper-parameter, if $c_{l,j}>t'$, we mark $x_{l,j}$ as a layer-wise ``cancer'' feature map and perform the same feature rescaling on it to remove image artifacts.
As Fig.~\ref{fig:ablation} shows, this layer-wise variant of our method is also effective for certain types of image artifacts (\eg, background intrusion).
While for some other types of artifacts (\eg, alien objects), it becomes less effective (Fig.~\ref{fig:Layer_wise}).

\begin{figure}[t]
  \centering
  \includegraphics [width=0.48\textwidth]{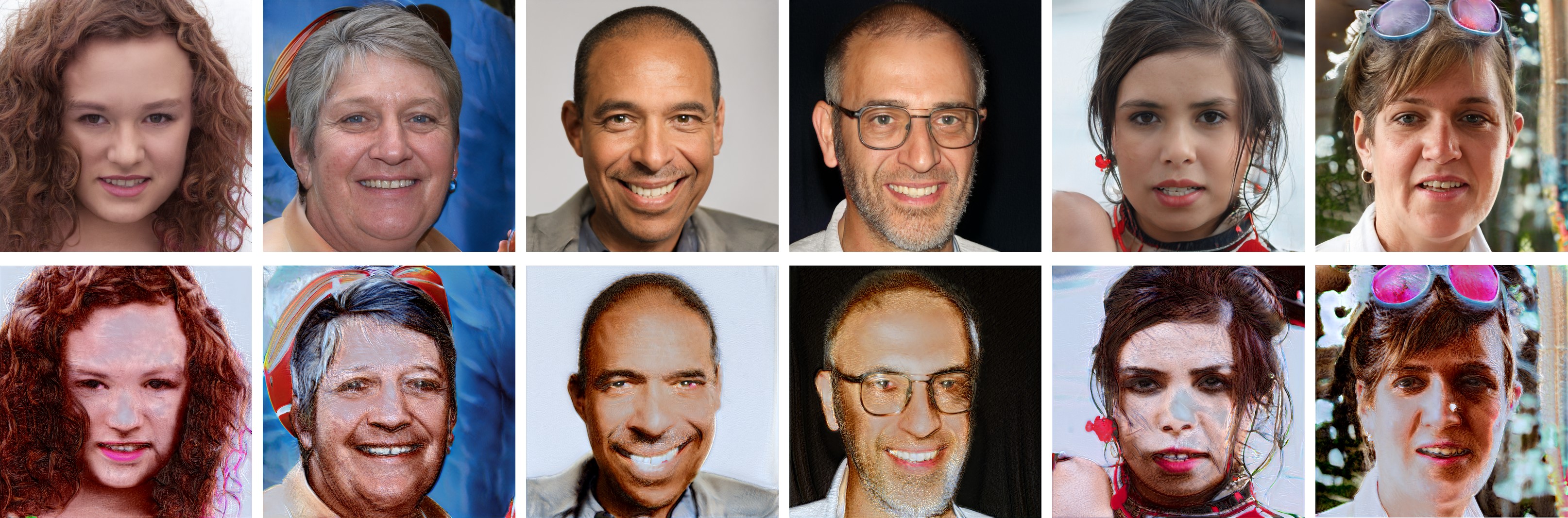}
    \caption{Effects of pixel-wise feature vector normalization on StyleGAN2 pre-trained on FFHQ dataset. Top row: normal images synthesized by StyleGAN2. Bottom row: images after applying pixel-wise feature vector normalization, which become unrealistic.}
  \label{fig:Pixel-wise normalization}
\end{figure}

\vspace{1mm}
\noindent \textbf{Pixel-wise ``Cancer'' Curing.}
We follow ProGAN~\cite{karras2017progressive} and apply its {\it pixel-wise feature vector normalization} technique, which was originally proposed to stabilize GAN training by resolving the potentially exploding magnitudes in the generator and discriminator, to pre-trained StyleGAN2.
However, we observed that even the normally synthesized images become unrealistic (Fig.~\ref{fig:Pixel-wise normalization}), indicating that it is not applicable to our problem.

\begin{table}[t]
  \caption{Time costs of generating a single image for StyleGAN2, Ours (Serial) and Ours (Parallel) on an Nvidia V100 GPU. The time is averaged over 100 runs.}
  \vspace{2mm}
  \label{tb:time complexity}
    \centering
    \begin{tabular}{lrr}
        \toprule
        Dataset
        \centering
         &\multicolumn{1}{c}{FFHQ} &\multicolumn{1}{c}{AFHQ}                     \\
        \midrule
        SG2 & 0.075s  & 0.031s  \\
        Ours (Serial) & 0.972s & 0.721s  \\
        Ours (Parallel) & 0.409s & 0.189s  \\
        \bottomrule
    \end{tabular}
    \vspace{-4mm}
\end{table}

\begin{figure*}[t]
  \centering
  \includegraphics [width=0.99\linewidth]{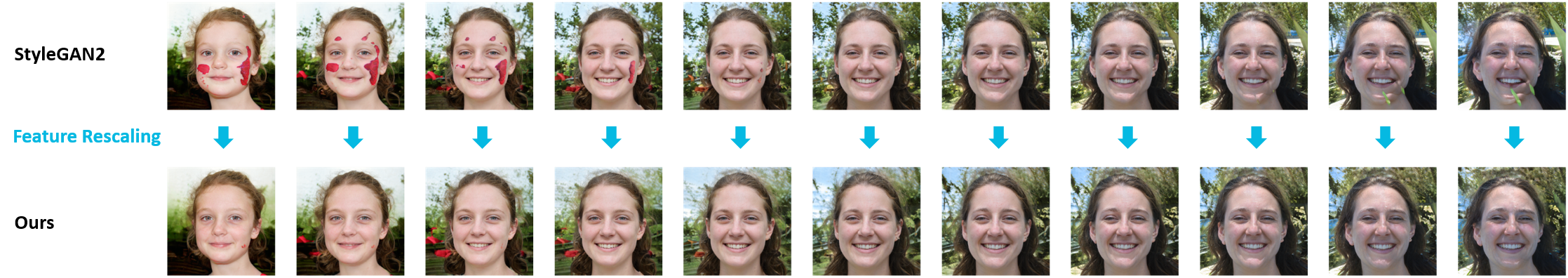}
  \vspace{2mm}
  \caption{Our method is compatible with StyleGAN interpolations. StyleGAN2: interpolation between defective images synthesized by StyleGAN2; Ours: images ``corrected'' by our feature rescaling method.}
  \label{fig:Interpolation}
\end{figure*}

\begin{figure*}[t]
  \centering
  \includegraphics [width=0.99\linewidth]{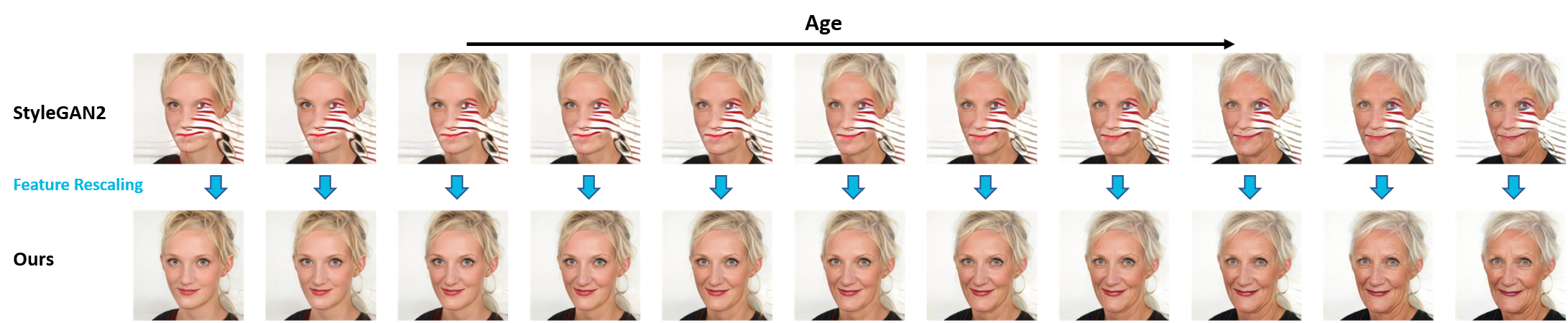}
  \vspace{2mm}
  \caption{Our method is compatible with StyleGAN image editing~\cite{viazovetskyi2020stylegan2}. StyleGAN2: interpolation between defective images synthesized by StyleGAN2; Ours: images ``corrected'' by our feature rescaling method.}
  \label{fig:Editing}
\end{figure*}

\subsection{Time Complexity}
Table~\ref{tb:time complexity} shows the time costs of generating a single image for StyleGAN2, ours implemented in serial, and ours implemented in parallel on an Nvidia V100 GPU. It can be observed that the parallel implementation of our method significantly reduces the time costs from a factor of approximately 15-20 (Serial) to approximately 5 times those of the original StyleGAN2 (SG2). We leave further acceleration of our method to future work.

\subsection{Applications in Interpolation and Editing}

Fig.~\ref{fig:Interpolation} shows that our method can remove image artifacts while retaining smooth StyleGAN interpolations, and Fig.~\ref{fig:Editing} shows similar observations for StyleGAN image editing. 
These suggest that our method preserves StyleGAN latent semantics and is thus compatible with various StyleGAN latent space operations.

\section{Conclusion}

In this paper, we propose a novel feature identification and rescaling method to address the artifacts in StyleGAN synthesized images. 
The rationale of the proposed feature rescaling method stems from the {\it Feature Proliferation} phenomenon we discovered that has a strong causal relationship with StyleGAN image artifacts.
Specifically, we observed that the features that deviate significantly from the mean of their distribution are highly likely to dominate the output of a neuron and proliferate in the forward propagation.
As a result, the output of the entire network will be dominated by a small number of features.
We ascribe this to the strong statistical assumption used in the weight modulation and demodulation used by the StyleGAN family.
Intuitively, we name Feature Proliferation as the ``cancer'' in StyleGANs from its proliferating and malignant nature.
Experimental results demonstrate that our method achieves a better trade-off between quality and diversity of StyleGAN synthesized images than the popular truncation trick.

\vspace{2mm}
\noindent \textbf{Limitation and Future Work.}
Although effective and significantly better than the truncation trick, the proposed method is not perfect as it can still remove some useful image features and make minor changes to high-quality StyleGAN synthesized images. We plan to design more precise methods to identify and process proliferating features in future work.
In addition, we believe that Feature Proliferation is not a problem exclusive to StyleGAN and may occur in other works as long as weight modulation and demodulation techniques are used (Sec.~\ref{sec:root_cause}). Thus, we plan to investigate in future work whether feature proliferation is a common phenomenon in various network architectures and its consequences in different tasks.

\vspace{2mm}
\noindent \textbf{Potential Negative Social Impact.}
Although our method works as a post-hoc ``cure'' for StyleGAN synthesized images and can save millions of dollars in training costs and energy consumption, it may also lead to potentially negative social impacts.
For example, it can further improve the quality of StyleGAN synthesized images, which may facilitate the creation of fake portrait photos/videos for fake social media accounts and fake news. 
Nevertheless, we still believe that our work will make a positive contribution to the field and provide new insights into the mechanisms of generative models that may help identify such fake content.

\section*{Acknowledgements}
\noindent
We thank the reviewers for their constructive comments that helped improve this paper.
This work was partially supported by the China Scholarship Council (CSC) under Grant No.~202106150033 and the UK Engineering and Physical Sciences Research Council (EPSRC) through the Doctoral Training Partnerships (DTP) with No. EP/T517951/1 (2599521).

{\small
\bibliographystyle{ieee_fullname}
\bibliography{egbib}
}


\appendix

\twocolumn[
\begin{@twocolumnfalse}
    \begingroup
        \newpage
        \null
        \vskip .375in
        \begin{center}
      {\Large \bf Feature Proliferation --- the ``Cancer'' in StyleGAN and its Treatments\\Supplementary Materials \par}
      
        \end{center}
    \endgroup
    \vskip .5em
    \vspace*{12pt}
\end{@twocolumnfalse}
]

\section{\hbox{Additional Comparisons of Difference Maps}}
\begin{figure}[!h]
  \centering
  \subfigure[Cat]{
  \begin{minipage}[t]{1\linewidth}
  \centering
  \includegraphics[width=0.65\textwidth]{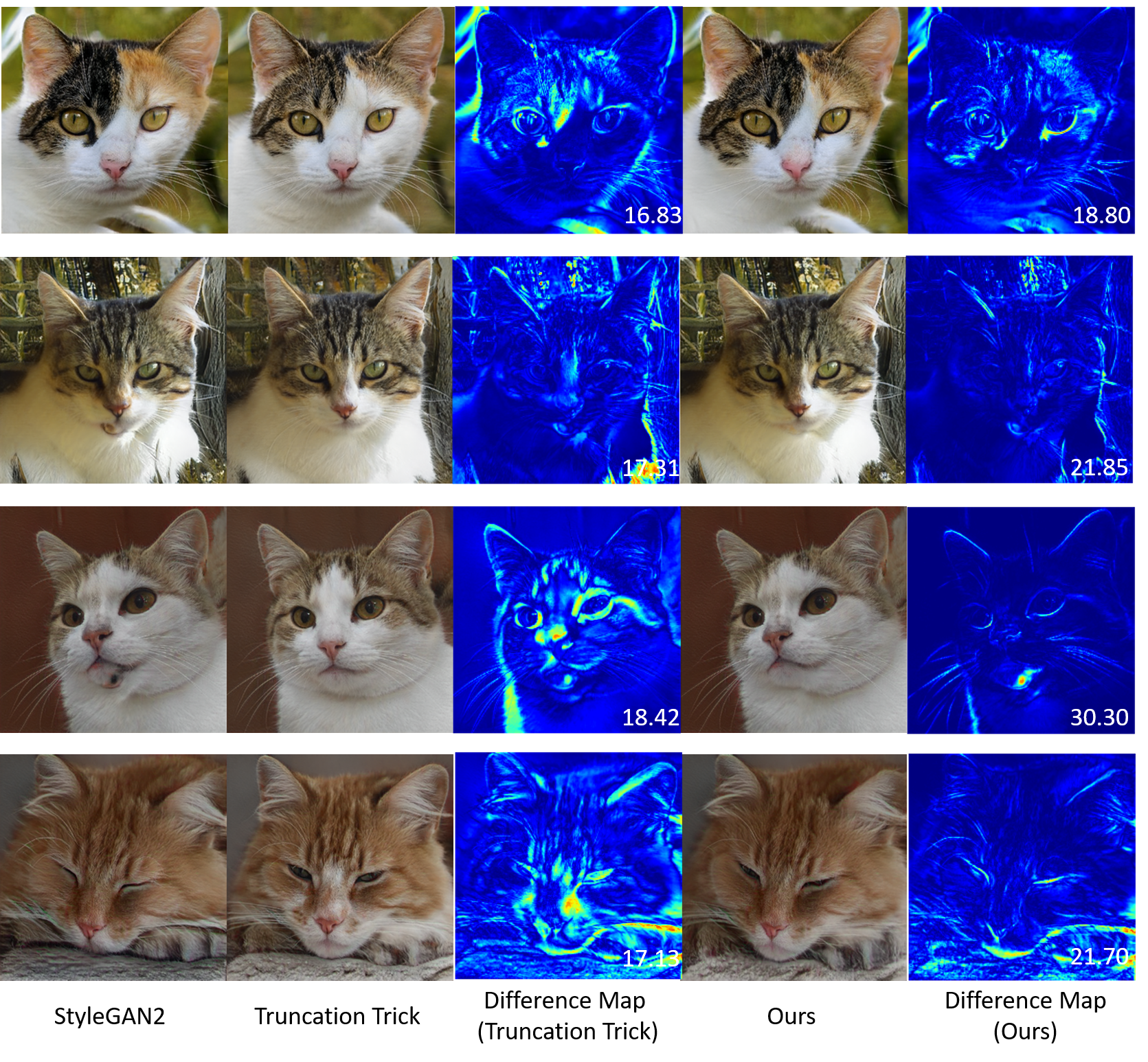}\label{Difference_map_cat}
  \end{minipage}%
  }
  
  \subfigure[Metface]{
  \begin{minipage}[t]{1\linewidth}
  \centering
  \includegraphics[width=0.65
  \textwidth]{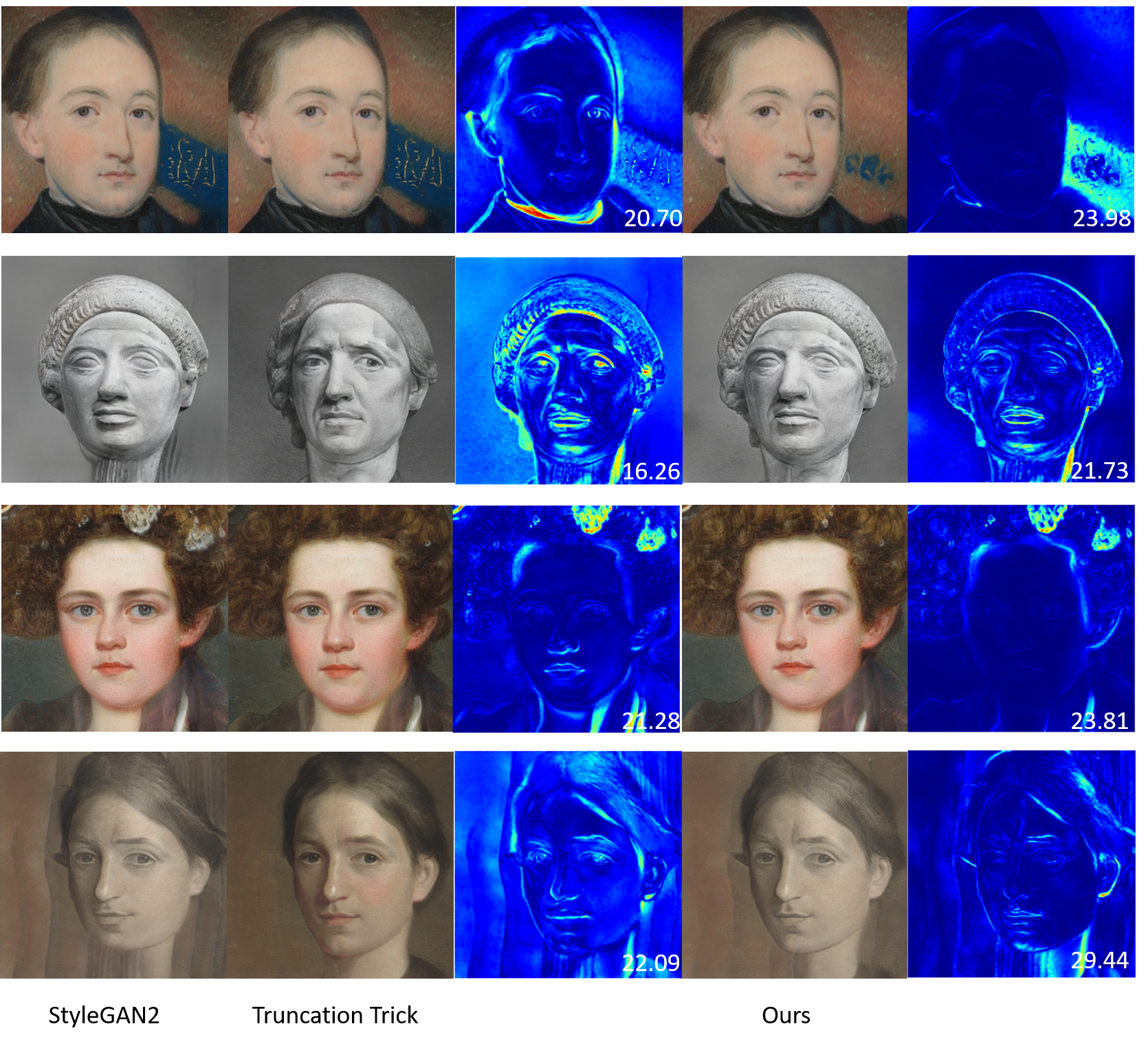}\label{Difference_map_metface}
  \end{minipage}%
  }    
  \caption{Comparison of difference maps of truncation trick and our method over (a) AFHQ-Cat~\cite{choi2020stargan} and (b) MetFace~\cite{karras2020training} datasets. The numbers at the  bottom right corner of the difference images are PSNR scores.}
  \label{fig:difference_map_supp}
\end{figure}

Fig.~\ref{fig:difference_map_supp} shows additional comparisons of difference maps and PSNR scores of the truncation trick and our method on Cat and MetFace datasets. Our method is less destructive and better retains useful features in the generation.

\section{Additional Statistics of Dominant Features} 

\begin{figure}[h]
  \centering
  \vspace{-1mm}
  \includegraphics [width=0.98\linewidth]{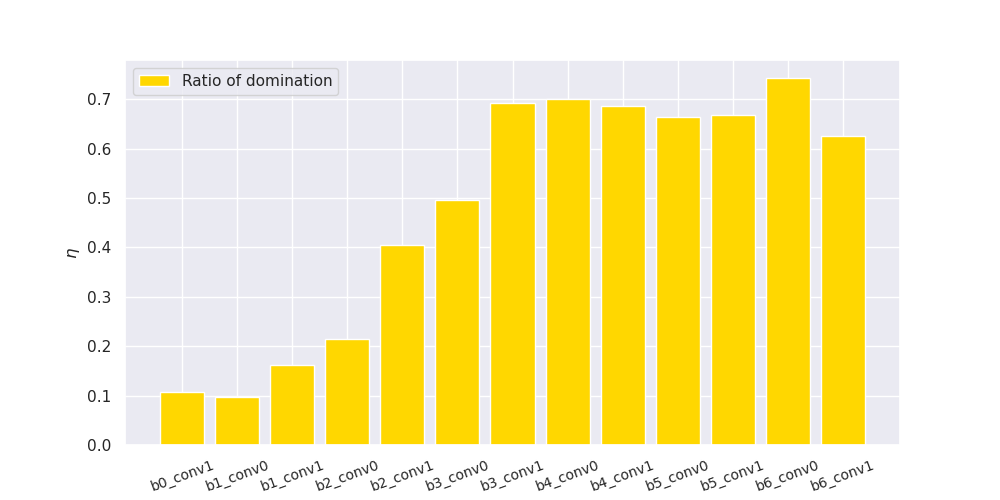}
  \caption{The ratio of dominant features $\eta$ increases in different layers (from left to right) in a representative StyleGAN2 synthesized image affected by ``cancer''. 
  }

  \vspace{-2mm}
  \label{fig:domination_layer-wise}
\end{figure}

In addition to the statistics of dominant features in Fig.~4 of the main paper (\ie, the $\eta$ in \textcolor{red}{red}), we include more fine-grained results in Fig.~\ref{fig:domination_layer-wise}, showing more details on how the ratio of dominant features increase across layers.

\section{Results on Non-Face Data} 

\begin{figure}[h]
  \centering
  \vspace{-1mm}
  \includegraphics[width=0.98\linewidth]{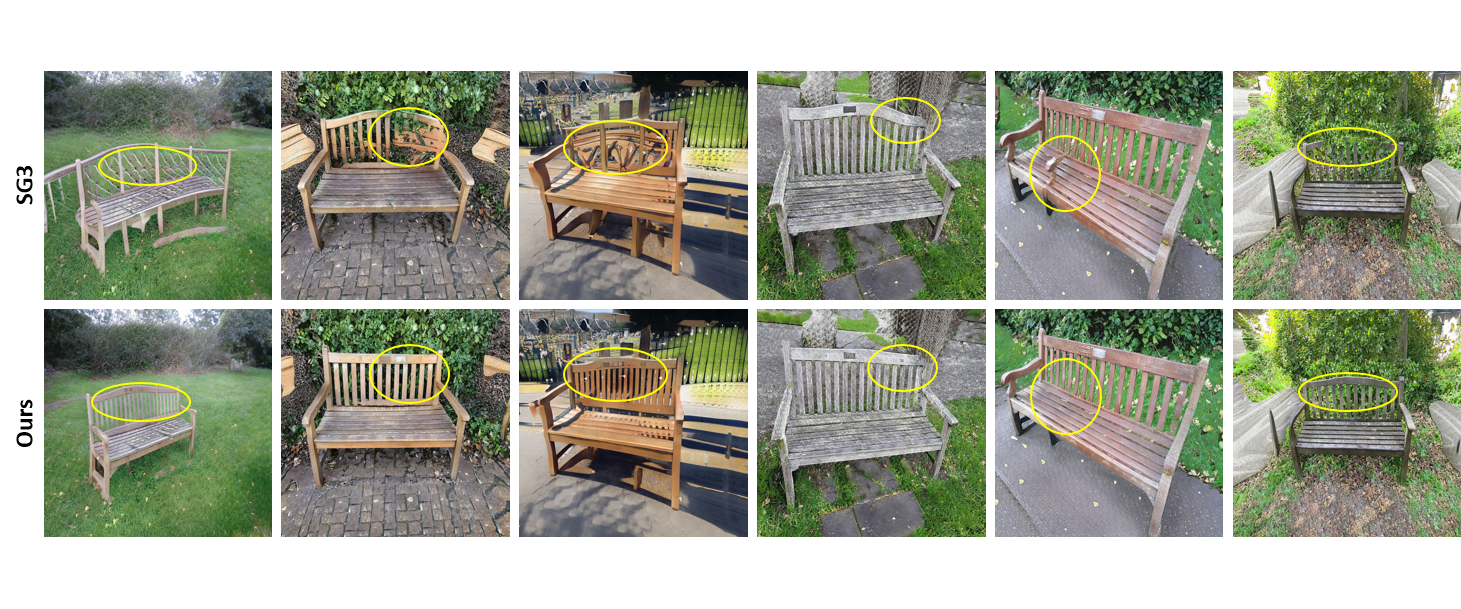}
  \vspace{-1mm}
  \caption{Results on the Bench dataset. SG3: StyleGAN3.}
  \vspace{-3mm}
  \label{fig:BENCH}
\end{figure}

Fig.~\ref{fig:BENCH} shows that our method works well on the Bench Dataset\footnote{\url{https://github.com/justinpinkney/awesome-pretrained-stylegan3}} without ``face'' structures.

\section{Choice of Hyper-parameters (Cont.)}

\begin{figure*}
  \centering
  
  \subfigure[]{
  \begin{minipage}[t]{0.75\linewidth}
  \centering
  \includegraphics[width=0.99\textwidth]{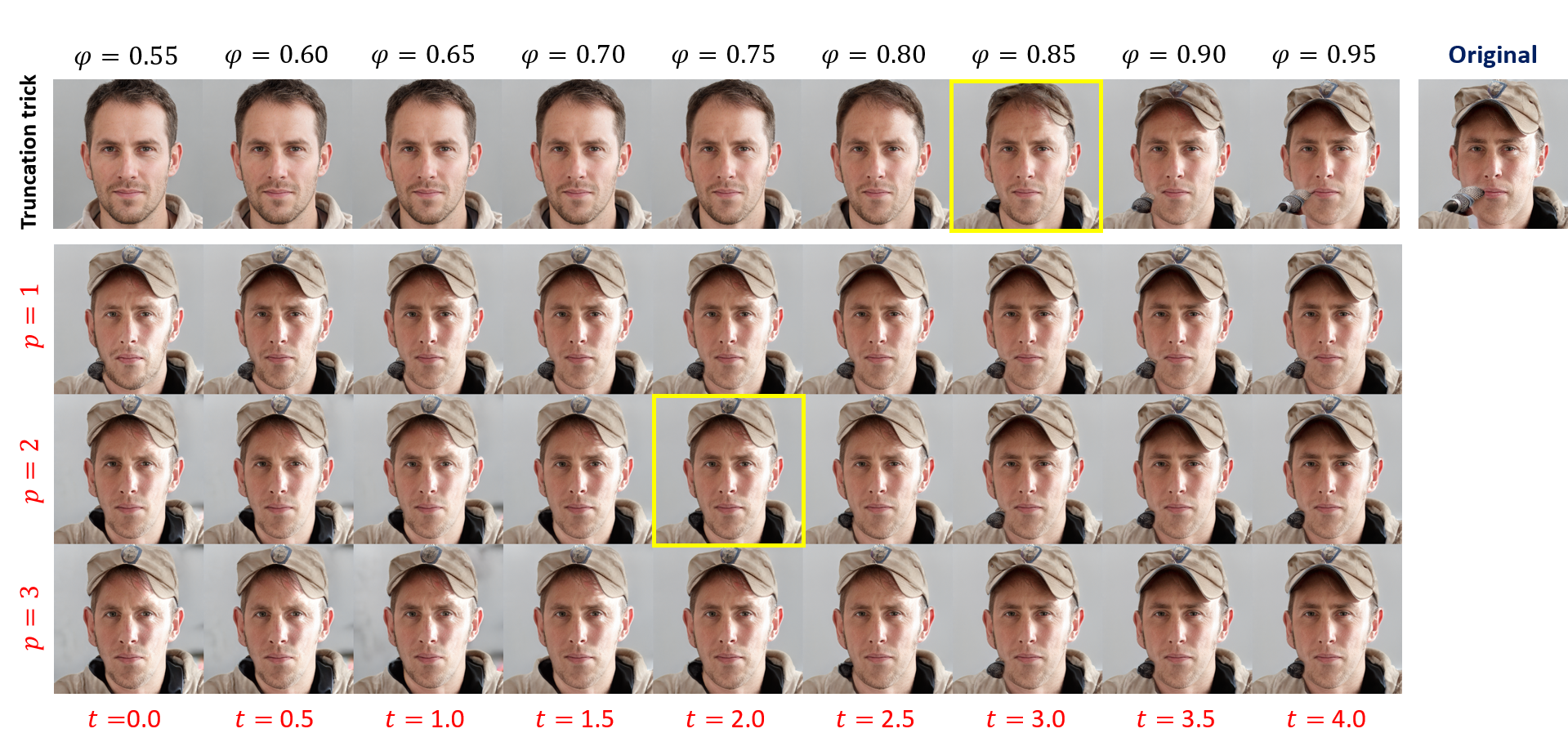}\label{hyper_t_p_2}
  \end{minipage}%
  }
  \subfigure[]{
  \begin{minipage}[t]{0.75\linewidth}
  \centering
  \includegraphics[width=0.99\textwidth]{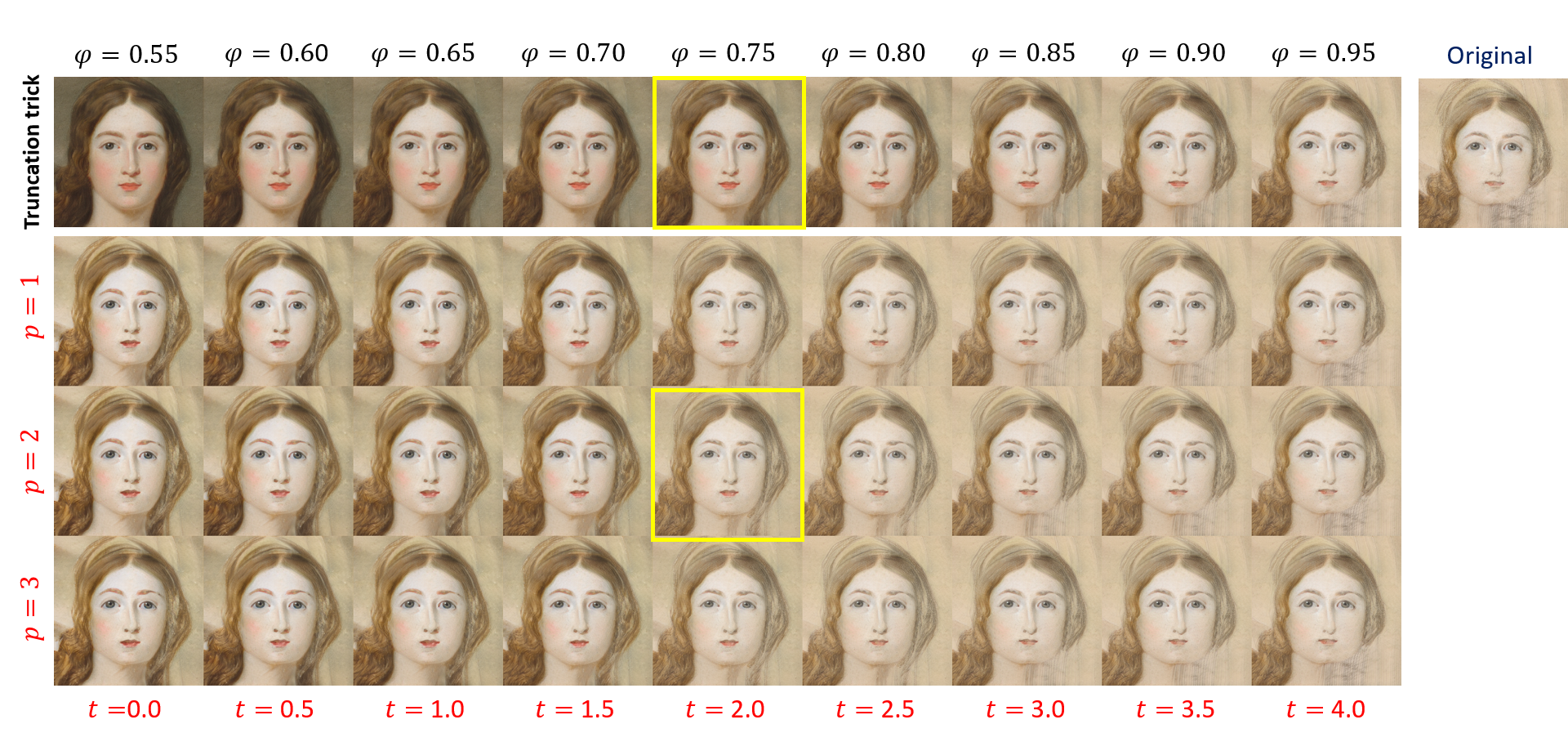}\label{hyper_t_p_1_supple}
\end{minipage}%
  } 
  \subfigure[]{
  \begin{minipage}[t]{0.75\linewidth}
  \centering
  \includegraphics[width=0.99\textwidth]{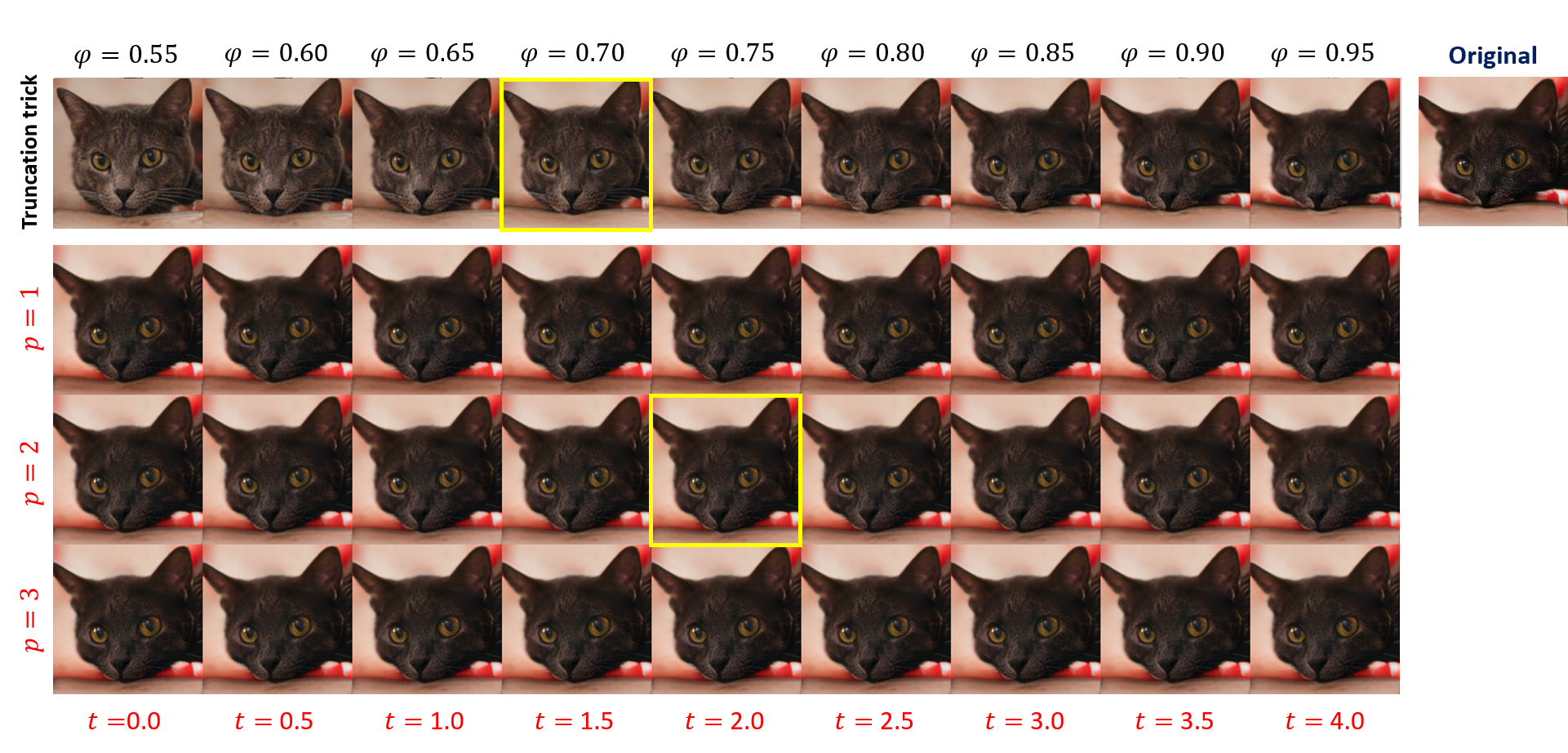}\label{add_hyper_t_p_2}
  \end{minipage}%
}
\caption{Choice of hyper-parameters (Cont.). (a) FFHQ~\cite{karras2019style}; (b) MetFace~\cite{karras2020training}; and (c) AFHQ-Cat~\cite{choi2020stargan} datasets.
For each subfigure, Top row: images generated with truncation trick with $\psi=0.55$ to $0.95$; Rows 2-4: images generated with our method with $t=0.0$ to $4.0$, $p=1$ to $3$. Our method removes the artifacts while retaining almost all important features of the original image, indicating that it achieves a better trade-off between image quality and diversity than the truncation trick.}
  \label{hyper_supple}
\end{figure*}

\vspace{-2mm}
In addition to Fig.~8 in Sec.~5.4 of the main paper, we provide additional qualitative justification for our choice of $p=2, t=2$ with FFHQ~\cite{karras2019style}, MetFace~\cite{karras2020training} and AFHQ-Cat~\cite{choi2020stargan} datasets. As Fig.~\ref{hyper_supple} shows, our choice of hyper-parameters is valid across different datasets.

%

\section{Additional Results for Applications in Interpolation and Editing}

\begin{figure*}
  \centering
  \includegraphics [width=0.98\linewidth]{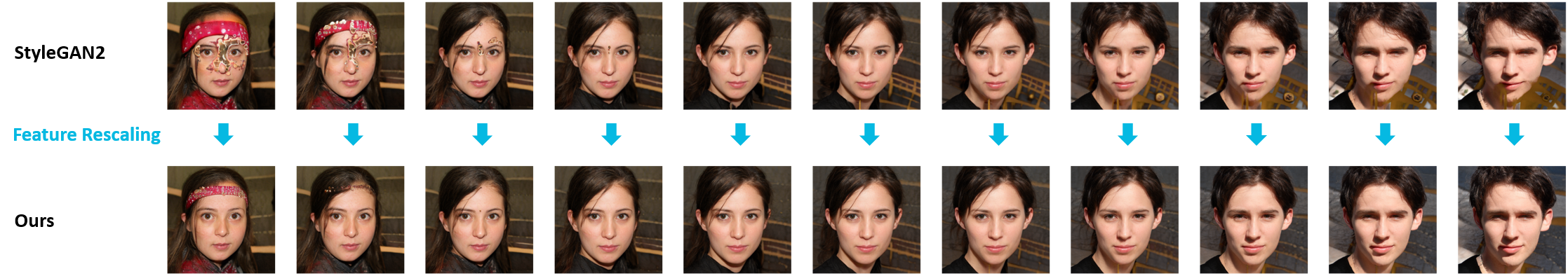}
  \vspace{1mm}
  \caption{Our method is compatible with StyleGAN interpolations. StyleGAN2: interpolation between defective images synthesized by StyleGAN2; Ours: images ``corrected'' by our feature rescaling method.}
  \vspace{0mm}
  \label{fig:Interpolation_supple}
\end{figure*}

\begin{figure*}
  \centering
  \includegraphics [width=0.98\linewidth]{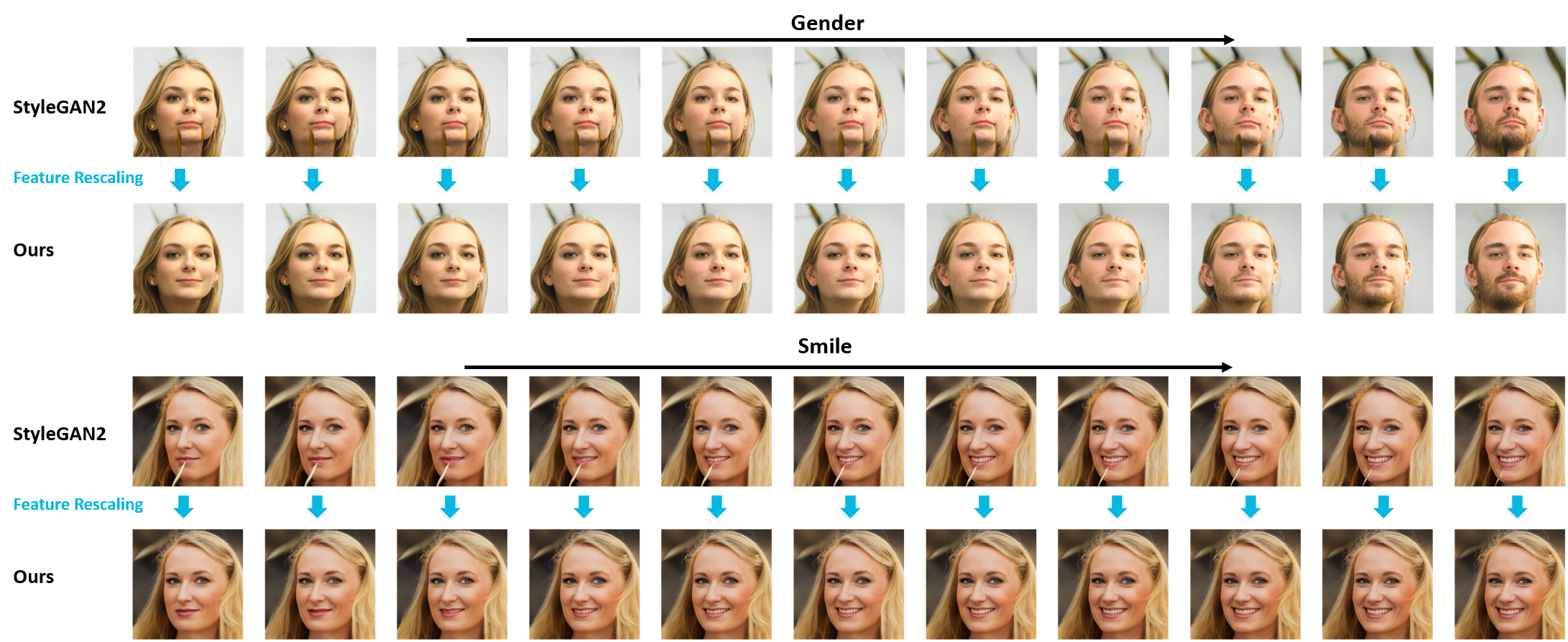}
  \vspace{1mm}
  \caption{Our method is compatible with StyleGAN image editing~\cite{viazovetskyi2020stylegan2}, \eg, gender, and smile. StyleGAN2: interpolation between defective images synthesized by StyleGAN2; Ours: images ``corrected'' by our feature rescaling method.}
  \vspace{0mm}
  \label{fig:Editing_supple}
\end{figure*}

Fig.~\ref{fig:Interpolation_supple} shows additional results for the application of our method in StyleGAN image interpolation. Our method can still remove image artifacts while retaining smooth StyleGAN interpolations. Fig.~\ref{fig:Editing_supple} shows similar observations for StyleGAN image editing. 
These results demonstrate that our method retains StyleGAN latent semantics and is compatible with various StyleGAN latent space applications.

\end{document}